\documentclass[mnsc]{informs3}

\OneAndAHalfSpacedXI 
\TheoremsNumberedThrough    
\ECRepeatTheorems

\EquationsNumberedThrough  
\usepackage{natbib}
 \bibpunct[, ]{(}{)}{,}{a}{}{,}%
 \def\bibfont{\small}%
\usepackage{listings}
\usepackage{xcolor}
\usepackage{caption}      % for \captionof
\usepackage{caption}
\usepackage{threeparttable}
\usepackage{subcaption}
\usepackage{palatino}
\usepackage{mathpazo}
\usepackage{booktabs}
\usepackage{changepage}
\usepackage{todonotes}
\usepackage{adjustbox}
\usetikzlibrary{calc}
\definecolor{FollowUpGray}{gray}{0.50}

\usetikzlibrary{positioning,fit,backgrounds,arrows.meta}
\usepackage{tikz}
\usetikzlibrary{positioning, fit, backgrounds, arrows.meta}
\makeatletter
\def\theARTICLETOPLEFT{}
\def\theARTICLETOPRIGHT{}
\def\theARTICLETOP{}
\makeatother

\newcommand{\blfootnote}[1]{%
  \begingroup
  \renewcommand\thefootnote{}\footnote{#1}%
  \addtocounter{footnote}{-1}%
  \endgroup
}

\begin{document}
%%%%%%%%%%%%%%%%

%\RUNTITLE{Better Together: Quantifying the Benefits of Human-AI Collaboration in Interviewing}

% Clear running headers
\renewcommand{\RRHFirstLine}{}
\renewcommand{\RRHSecondLine}{}
\renewcommand{\LRHFirstLine}{}
\renewcommand{\LRHSecondLine}{}

\renewcommand{\ECRRHFirstLine}{}
\renewcommand{\ECRRHSecondLine}{}
\renewcommand{\ECLRHFirstLine}{}
\renewcommand{\ECLRHSecondLine}{}

\def\setoddRH{}
\def\setevenRH{}
\def\setoddECRH{}
\def\setevenECRH{}
\setlength{\footskip}{20pt}  
\pagestyle{plain}  

\TITLE{Quantifying the Benefits of AI-Assisted Recruitment}

\ARTICLEAUTHORS{%

\begin{minipage}[t]{0.48\textwidth}
\AUTHOR{Ada Aka*}
\AFF{Stanford University, \EMAIL{adaaka@stanford.edu}}
\end{minipage}
\hfill
\begin{minipage}[t]{0.48\textwidth}
\AUTHOR{Emil Palikot*}
\AFF{Stanford University, \EMAIL{palikot@stanford.edu}}
\end{minipage}

\vspace{0.5em}

\begin{minipage}[t]{0.48\textwidth}
\AUTHOR{Ali Ansari}
\AFF{Stanford University \& micro1, \EMAIL{aliansarinik@stanford.edu}}
\end{minipage}
\hfill
\begin{minipage}[t]{0.48\textwidth}
\AUTHOR{Nima Yazdani}
\AFF{University of Southern California \& micro1, \EMAIL{nimayazd@usc.edu}}
\end{minipage}

}
\ABSTRACT{

Hiring algorithms have mostly scored the materials recruiters already see. Large language models (LLMs) can instead generate new information about candidates by conducting, at scale, structured interviews once reserved for a few finalists. We study this shift in two field experiments at a recruitment platform. The first experiment holds the candidate pool fixed and randomizes whether recruiters observe the AI Interview Report; the second embeds the AI interview as a requirement in a live hiring pipeline. In both, candidates shortlisted with AI interview information pass the final human interview (conducted blind to shortlisting condition) at rates 17.5 (SE 8.5) to 20 (SE 11.8) percentage points higher than candidates shortlisted from resumes alone. The gains concentrate where resumes are least informative: adding AI Interview Report ratings to conventional candidate features raises out-of-sample AUC by 0.18 for junior candidates, against 0.08 for non-junior candidates. The participation cost falls on applicants as 75 percent of invited candidates do not complete the interview. However, the attrition is itself a signal: completion is more consistent with job-search motivation than with predicted interview performance. AI interviews thus add information exactly where conventional signals fail, and they move the cost of screening from firms to applicants.

}


%\ACKNOWLEDGMENT{We are grateful to the editor, the associate editor, and the anonymous reviewers for their very constructive guidance. 

%%%% REMAINING TO-DOs: 
%Correct the Experiment 2 estimand language and equation so ITT is truly defined at assignment, not conditional on reaching finals.
%Add a very explicit step-by-step funnel description for Experiment 2, with exact counts, exact decision rules, and exactly what humans saw at each stage.
%Resolve the blindness issue head-on.
%Remove all leftover placeholders and any rounding inconsistencies.
%Tone down broad claims from “AI improves hiring quality” to the narrower claims your design actually identifies, especially in Experiment 2.
%All invited revisions must adhere to the following page limit: 47 pages with 25 lines of text per page (double spacing) or 32 pages with 33 lines of text per page (1.5 spacing). An online appendix will not count toward the page limit.
%ILL CHECK ACCURACY ETC AGAIN
%Need to make sure condition names are consistnet

%%%%%%%
\maketitle
\blfootnote{We thank Nir Halevy, Paul Oyer, and participants at the 2025 Columbia AIML Conference, CODE@MIT, and the Marketing Science Conference for their helpful comments.}
\newpage
\section{Introduction}
The top of the hiring funnel increasingly combines scale with weak signals. Large employers process hundreds of thousands of applications a year \citep{horton2021job, horton2024reducing, Brownstein2024GoldmanInternship, birinci2025job}, while relying on application materials that have become less informative. One reason is that applicants increasingly use large language models (LLMs) to draft and reformat resumes, potentially homogenizing keyword density and phrasing across candidates \citep{doshi2024generative, nejad2025labor, cui2025signaling,cowgill2026does}. Applicants also exaggerate: 78\% of job seekers report having considered overstating their skills and 60\% admit to claiming in-demand skills they lack \citep{henle2019assessing, ResumeLab2023, freedman2024people}. Faced with these pressures, more than 90\% of employers have adopted some form of AI-assisted hiring \citep{fuller2021hidden, Mujtaba}. We ask what this adoption delivers in practice, and through which channels.

A structured AI interview is an AI-led assessment in which candidates answer a standardized sequence of job-relevant questions, with adaptive follow-ups and an AI-generated report summarizing their skills for the recruiter (i.e., an \emph{AI Interview Report}). Deploying such an interview can affect hiring on two distinct margins. First, the interview can give the recruiter information about a candidate that resumes cannot easily capture, improving selection from a fixed applicant pool. Second, when the interview is required at application, it imposes a participation cost on applicants and reshapes who remains in the pool. These margins identify different objects: a recruiter who reads an \emph{AI Interview Report} sees a richer candidate, while a candidate who completes an AI interview might reveal a stronger preference for the job.  We causally identify the report’s informational value among interview completers and separately characterize the participation effects of requiring the interview.

We run two randomized experiments on the same recruitment platform. Experiment 1 isolates the information channel by holding the candidate pool fixed and varying only what the recruiter sees. To do that we advertised four job positions, received 4,323 applications, and invited all of them for an AI interview.\footnote{In both experiments, advertised job positions were real job openings leading to actual employment.} 1,108 candidates completed it, generating an \emph{AI Interview Report} on each of them. We then randomized each candidate to one of two recruiter-types. The control recruiter sees the candidate's resume and \emph{Resume Score}, a machine learning generated metric of the candidate's match for the position.  The treatment recruiter sees the resume, the \emph{Resume Score}, and the \emph{AI Interview Report}. Recruiters are drawn from a pool of recruiters available on the platform and remain in the same treatment condition throughout the experiment. Each recruiter independently selects a shortlist for the final round. Shortlisted candidates from both arms are then interviewed by the same hiring manager, who is blind to the experimental arm and makes the pass/fail decision, which we use as the primary outcome. Because every candidate completed the AI interview before assignment, self-selection into the interview cannot drive the contrast between arms; only the recruiter's information set varies. Treatment recruiters shortlist candidates who pass the final blind interview at 46\% compared with 29\% under resume screening alone, a 17.5-percentage-point difference in pass rates (SE 8.5~pp). Notably, treatment recruiters select candidates with weaker \emph{Resume Scores} but stronger AI-assessed skills, so the gap \emph{widens} once we hold the resume fixed: the \emph{AI Interview Report} does work the resume cannot.

Experiment 2 adds the AI Interview to a regular hiring pipeline. We advertised a Junior Frontend Engineer position and received 34,493 applications, which were randomized at the moment of application. Control applicants followed a conventional screening process, without the AI interview, in which the application materials, including a \emph{Resume Score}, are forwarded to the recruiter. Treatment applicants were first invited to complete the AI interview, and the resulting \emph{AI Interview Report} (for the candidates who complete it), together with the resume and \emph{Resume Score}, are passed to recruiter review. As the interview takes 30-40 minutes to complete, only approximately 25\% of invited applicants complete it (a similar ratio of completed to invited as in Experiment 1). Recruiters and the hiring manager work in an analogous way to Experiment 1. Control group recruiters shortlist candidates based on the resume and \emph{Resume Score}, while treatment group recruiters in addition see the \emph{AI Interview Report}. Hiring managers are blind to the treatment condition, conduct an interview, and make a pass or fail decision. Restricting attention to candidates who were shortlisted by a recruiter for the final interview, treatment-arm finalists pass at a rate 20.0 percentage points higher than control-arm finalists (SE 11.8).

Behind these reduced-form effects we document two mechanisms. The first is skill verification. The AI interview provides recruiters with evidence about whether candidates can demonstrate skills that appear on their resumes. We find that about 21\% of treatment-arm completers in Experiment~2 list at least one required technical skill (i.e., React, JavaScript, or CSS) that the AI rates as ``Not Familiar,'' (the lowest possible grade representing lack of any demonstrated proficiency). A separate audit of 720 candidates drawn from the platform's historical data returns a near-identical 21.6\%. To validate the AI's skill ratings against an external benchmark, an independent expert scored a random subset of 60 recorded AI interviews. The expert was asked to provide a binary classification of whether the candidate had demonstrated any proficiency in the evaluated skill, which we then compared to whether the AI graded the candidate as ``Not Familiar''. The expert and the AI agreed on 96.1\% of skill classifications, providing convergent validity for the ratings that drive the information channel. 

The second mechanism is selective participation: candidates who drop out of the AI interview are not a random subset of those invited. Observable characteristics such as resume quality, education, experience, and gender predict dropout but jointly explain less than one percent of its variance, suggesting that unobservables such as motivation, outside offers, or beliefs about AI evaluation drive most of the variation. Five-month LinkedIn data are consistent with this reading. Among shortlisted candidates from the treatment group, 41\% report a new job five months later, against 21\% for the treatment group candidates who completed the AI interview and failed it, 23\% for those who completed and passed, and 16\% among the candidates who did not complete the interview. This suggests that completing the AI interview is indicative of the candidate's job search motivation.

Finally, we use historical platform data to ask where the \emph{AI Interview Report} adds the most predictive content. Because our experiments focus on junior technical roles, we examine whether the report is especially valuable in that segment or similarly informative across role types. We use a historical sample of 4,372 candidates across 183 roles who completed both an AI interview and a final human interview, train gradient-boosted classifiers on half of the sample, and evaluate predictive performance on the held-out half. The baseline model uses conventional candidate and role features; the AI-augmented model additionally includes \emph{AI Interview Report}. Adding AI interview information raises out-of-sample AUC for predicting final-interview success from 0.62 to 0.79 for junior candidates ($\Delta = 0.18$), compared to a lift from 0.78 to 0.86 ($\Delta = 0.08$) in the non-junior positions sample. The \emph{AI Interview Report} is most informative exactly where resumes are least informative: among candidates with shorter track records and weaker conventional signals.

Two limitations bound the reading. First, both experiments cover junior technical roles, and our predictive analysis suggests that this is the segment where AI interview assessments add the most. We therefore read our experimental magnitudes as identifying the upper end of what an AI interview can deliver, not an average across the labor market. Second, we do not observe long-run job performance. Our available outcomes are interview pass rates and a five-month LinkedIn employment proxy, but not post-hire retention or productivity. In the following section, we review the related literature that our work connects to.

\subsection{Related Literature}
\label{sec:lit}

Our paper relates primarily to two literatures: algorithmic evaluation of job candidates and the procedural design of hiring pipelines. The closest precursors to our work study algorithmic screening of conventional application materials. \cite{Hoffman2017} show that managers who override an algorithmic job-test score hire workers with shorter tenures than the algorithm would have selected; \cite{cowgill2020bias} shows that an ML resume screener improves selection relative to human reviewers and raises interview rates for nontraditional candidates, and \cite{autor2008does} find no evidence that standardized assessments harm minority applicants. These designs share a common structure. The algorithm operates on the same input class the recruiter would otherwise see, namely the resume, and either substitutes for or constrains human discretion over a fixed information set. Our studies, in contrast, focus on what new information about candidates AI can generate beyond the resume, and on the participation consequences of embedding such an AI process in the hiring pipeline.

A natural extension of this literature asks whether algorithms can evaluate interview content rather than resumes. \citet{naim2016automated} and \citet{hickman2022automated} analyzed verbal and non-verbal cues from automated video interviews at scale and showed that machine scoring of interview behavior can achieve reliability comparable to human raters. Large language models have since made it feasible to conduct and grade structured open-ended dialogue with a linguistic depth earlier rule-based systems could not reach \citep{Brown2020FewShot, Bommasani2021Foundation}. The closest comparisons to our setting are \cite{chakraborty2025can}, \cite{avery2026brave}, and \cite{jabarian2025voice}. \citet{chakraborty2025can} develop AI and AI-human models using recordings of conversational, two-sided video interviews, with a focus on detecting persuasion skills. \citet{avery2026brave} use AI and professional recruiters to evaluate the same pre-recorded, one-way audio and video interview responses. \citet{jabarian2025voice} run a field experiment with a similar underlying technology to our paper, an AI agent conducting structured interviews, in the customer-service segment of the labor market. Together, these studies examine the use of AI to conduct or evaluate interviews; we hold the candidate pool fixed and vary only the recruiter's information set, isolating the value of the \emph{AI Interview Report} itself, and separately estimate the effect of embedding that same interview in a live recruitment pipeline.
%The closest comparisons to our setting are \cite{chakraborty2025can}, \cite{avery2026brave}, and \cite{jabarian2025voice}. \citet{chakraborty2025can} and \citet{avery2026brave} use AI to evaluate pre-recorded, one-way interview responses. \citet{avery2026brave} also benchmark the AI evaluation against professional human evaluators of the same recordings. \citet{jabarian2025voice} run a field experiment with a similar underlying technology to our paper, an AI agent conducting structured interviews, in the customer-service segment of the labor market. These studies compare AI evaluation to human evaluation across separate candidate pools; we hold the candidate pool fixed and vary only the recruiter's information set, isolating the value of the \emph{AI Interview Report} itself, and separately estimate the effect of embedding that same interview in a live recruitment pipeline.

An important mechanism through which AI improves candidate selection is skill verification. Survey evidence finds that the majority of applicants admit to claiming in-demand skills they lack \citep{henle2019assessing, ResumeLab2023, freedman2024people}. A long tradition in personnel research finds that structured behavioral assessments are the most predictively valid selection device available to firms, outperforming resumes and unstructured interviews \citep{schmidt1998validity, campion1997review, levashina2014structured}, but their cost to employers limits their use to late-stage rounds with a handful of candidates. Cost is the binding constraint, and LLM capabilities change it by making behavioral assessment feasible at the top of the funnel, where verification may matter most. We contribute to this literature by showing that this is where the \emph{AI Interview Report}'s predictive content concentrates.

Another strand of literature studies how procedural requirements reshape who applies. \cite{koren2024gatekeeper} shows that introducing a screening gatekeeper changes both who applies and who completes the application, and \cite{hardy2017applicants} document that longer pre-hire assessments induce selective attrition. \cite{avery2026brave} randomize roughly 3{,}000 applicants for U.S.\ tech roles into an asynchronous AI-evaluated interview, a live online interview, or a no-interview control, and find that requiring the asynchronous AI-evaluated interview reduces application continuation by approximately 53 percent, with deterrence concentrated among the most qualified applicants and among women. We provide causal evidence on a related margin: imposing a 30-40~minute AI interview as a participation requirement at application yields a 75-percent dropout rate. Consistent with \citet{avery2026brave}, socio-demographic characteristics and role fit predict the decision to drop out, but most of the variation is unexplained by these observables. To characterize the unobserved component, we exploit an auxiliary outcome: whether the candidate reports a new job on LinkedIn five months later. Candidates who complete the AI interview, including those who fail it, are substantially more likely to report a new job than candidates who do not complete it. This pattern is consistent with selection on the intensity of job search and motivation to find a job, rather than on candidate ability or predicted interview performance alone.\footnote{Algorithm aversion could also contribute to the dropout pattern \citep{dietvorst2015algorithm, dietvorst2018overcoming, castelo2019taskdependent, liu2023algorithm}, particularly given recent evidence on demographic asymmetries in how applicants respond to AI-mediated evaluation \citep{an2024large, an2025measuring, salinas2023unequal, bogen2018help}.} Finally, the sample sizes of recruiter-advanced candidates in each experiment are too small to separately identify the contribution of selecting more motivated candidates from that of providing the recruiter with better information about those candidates. We treat the two channels as jointly operating rather than as separately identified.

Both literatures connect to the broader view of recruitment as a setting with asymmetric information, in which candidates can misrepresent their skills on resumes and firms must compensate with imperfect proxies \citep{akerlof1978market, spence1978job, tambe2019artificial}. Our contribution to this view is to characterize where in the candidate distribution AI-generated information adds the most.

The remainder of the paper proceeds as follows. Section~\ref{sec:setting} describes the recruitment platform, the two experiments, and the estimands. Section~\ref{sec:results} reports the main effects. Section~\ref{sec:mechanisms} examines the information and participation channels. Section~\ref{sec:auc} uses historical data to compare the value of the AI interview process across different position types. Section~\ref{sec:discussion} discusses external validity, implications for organizations adopting AI assessment, and directions for future research.
\section{Setting and Experimental Design}
\label{sec:setting}

This section describes the AI recruitment platform, the hiring pipeline in which both experiments are embedded, the two experiments and where each intervenes, and the variables we observe.

\subsection{The AI Recruiter Platform}
\label{sec:platform}

Our empirical analyses are conducted in the context of an LLM-powered interviewing system, AI Recruiter, developed by micro1 (an online recruitment platform). From the employer's job description, the platform generates a job-specific list of role-relevant skills and a corresponding interview blueprint. Candidates complete a 30-40 minute conversational interview structured on this blueprint, with a short coding task for technical roles. The interview is adaptive. An initial question is posed for each targeted skill, and the system generates follow-ups in real time, probing depth of understanding, practical application, and problem-solving approach. The transcript, together with the coding task output, constitutes the assessment record.

The platform produces an \emph{AI Interview Report} consisting of (i)~a per-skill proficiency rating on a four-point scale (\emph{Not Familiar}, \emph{Junior}, \emph{Mid-level}, \emph{Senior}); (ii)~a composite \emph{AI Interview Score} (0--100); (iii)~an automated proctoring score; and (iv)~a link to the interview video and full transcript (in practice, this link is most typically not used by recruiters). Skill ratings are generated by an LLM applied to the transcript and coding output, using the same evaluation approach for every candidate within a given role. The specific prompts and aggregation procedures are proprietary to micro1; we describe the system's inputs and outputs but do not have access to its internals. Section~\ref{sec:mechanisms} reports convergent-validity evidence comparing AI skill ratings against independent human-expert ratings of recorded AI interviews.

\subsection{The Hiring Pipeline}
\label{sec:hiring_process}
A conventional hiring pipeline can be represented in three steps: (i) a job position is advertised online and candidates apply; (ii) generalist recruiters shortlist candidates based on application materials together with a \emph{Resume Score} - a keyword based match score of a candidate and a position. The recruiter's task is to identify candidates who best match the job requirements, and there is no fixed decision rule or scoring rubric. The number of shortlisted candidates differs depending on the number of interview slots available in the next stage; (iii) a hiring manager, a domain expert often working in the team the candidate is getting hired for, interviews the candidates and makes a pass or fail decision (in practice in many cases there are several rounds of hiring manager interviews).

The AI interview sits within this hiring pipeline by adding an additional stage before the recruiters review the applications. Figure~\ref{fig:pipeline} illustrates this hiring pipeline with the AI interview. The pipeline has two natural intervention points. In Stage~2, where the AI interview is conducted (or not), and in Stage~3, where the \emph{AI Interview Report} enters (or does not enter) the recruiter's information set. Our two experiments intervene at these points separately.

% Add to the preamble:

%\usepackage{adjustbox}
\usetikzlibrary{positioning,arrows.meta}

\begin{figure}[t]
\centering
\caption{Experimental Pipeline}
\label{fig:pipeline}
\vspace{0.5em}

\begin{adjustbox}{max width=\textwidth,center}
\begin{tikzpicture}[
  font=\small,
  node distance=0.65cm,
  stage/.style={
    rectangle,
    rounded corners=3pt,
    draw=black!60,
    line width=0.6pt,
    minimum width=2.75cm,
    minimum height=1.5cm,
    inner xsep=6pt,
    align=center,
    fill=white
  },
  recruiterstage/.style={
    stage,
    minimum width=3.35cm,
    inner xsep=9pt
  },
  infobox/.style={
    rectangle,
    rounded corners=2pt,
    draw=black!40,
    line width=0.4pt,
    fill=black!3,
    inner sep=7pt,
    align=left,
    text width=6.2cm,
    font=\footnotesize
  },
  arrow/.style={
    -{Latex[length=2.5mm]},
    line width=0.6pt,
    draw=black!70
  },
  randomlabel/.style={
    rectangle,
    rounded corners=2pt,
    draw=black!70,
    line width=0.5pt,
    fill=black!5,
    inner sep=4pt,
    align=center,
    text width=3.7cm,
    font=\footnotesize\itshape
  },
  randomarrow/.style={
    dashed,
    -{Latex[length=2mm]},
    line width=0.5pt,
    draw=black!60
  }
]

% Stage nodes
\node[stage] (s1) {
Stage 1\\[2pt]
\textbf{Job Posting and}\\
\textbf{Application}\\[2pt]
\scriptsize Applicant submits resume
};

\node[stage, right=of s1] (s2) {
Stage 2\\[2pt]
\textbf{AI Interview}\\[2pt]
\scriptsize Conversational LLM interview;\\
\scriptsize \emph{AI Interview Report} generated
};

\node[recruiterstage, right=of s2] (s3) {
Stage 3\\[2pt]
\textbf{Human Recruiter}\\
\textbf{Shortlisting}\\[2pt]
\scriptsize Recruiter selects finalists
};

\node[stage, right=of s3] (s4) {
Stage 4\\[2pt]
\textbf{Human Hiring Manager}\\
\textbf{Interview}\\[2pt]
\scriptsize Blind to arm\\
\scriptsize Pass/Fail outcome
};

% Stage arrows
\draw[arrow] (s1) -- (s2);
\draw[arrow] (s2) -- (s3);
\draw[arrow] (s3) -- (s4);

% Information-set box under Stage 3
\node[infobox, below=0.75cm of s3] (info3) {%
\textbf{Recruiter sees:}\\[4pt]
\textbf{Control:} Resume + \emph{Resume Score}\\[4pt]
\textbf{Treatment:} Resume + \emph{Resume Score} + \emph{AI Interview Report}%
};

\draw[dotted, draw=black!50]
  (s3.south) -- (info3.north);

% Randomization callouts
\node[randomlabel, above=0.6cm of s2] (r2) {
\textbf{Experiment 2 randomization:}\\
Resume + \emph{Resume Score} only (control)\\
vs.\ \emph{AI Interview Report} (treatment)
};

\node[randomlabel, above=0.6cm of s3] (r3) {
\textbf{Experiment 1 randomization:}\\
Resume + \emph{Resume Score} only (control)\\
vs.\ \emph{AI Interview Report} (treatment)
};
\draw[randomarrow] (r2.south) -- (s2.north);
\draw[randomarrow] (r3.south) -- (s3.north);

\end{tikzpicture}
\end{adjustbox}

\vspace{6pt}

\begin{minipage}{0.95\textwidth}
\footnotesize
\textit{Notes:} The four-stage pipeline used in both experiments.
Experiment~2 randomizes at Stage~2 (whether the applicant is invited
to the AI interview before recruiter review). Experiment~1 randomizes
at Stage~3 (whether the recruiter sees the AI report when shortlisting).
The hiring manager at Stage~4 is blind to assignment in both experiments.
\end{minipage}

\end{figure}

%\paragraph{Stage 1: Posting and Application.} An employer posts a role on LinkedIn and the platform's website; applicants submit a resume.

%\paragraph{Stage 2: AI Interview.} The applicant is either invited to complete the AI interview before any human review, or routed directly to a recruiter. While Experiment~2 randomizes assignment between these two paths, in Experiment~1 all candidates complete the AI interview at this stage.

%\paragraph{Stage 3: Human Recruiter Shortlisting.} A recruiter reviews each candidate's materials and selects a shortlist for the final-round interview. In our experiments, the recruiter's information set is the only feature that varies across arms: control recruiters observe the resume and Resume Score, and treatment recruiters additionally observe the AI report. Both experiments intervene at this stage on the information available to the recruiter, but they differ in how that information is generated upstream.

%\paragraph{Stage 4: Human Hiring Manager Interview.} Shortlisted candidates meet a hiring manager for a structured final-round interview. The hiring manager observes the resume but is blind to the AI report and to treatment assignment, ruling out differential evaluator behavior as a channel for any observed treatment effect. A pass at this stage is the primary outcome in both experiments. In practice, passing is typically followed by additional hiring-manager rounds, but we treat the first such interview as the endpoint.

\subsection{Two Complementary Experiments}
\label{sec:two_experiments}

Experiment~1 randomizes the recruiter's information set at Stage~3: all candidates complete the AI interview, but only treatment-arm recruiters observe the resulting report. This isolates the informational value of the \emph{AI Interview Report} and allows us to study to what it adds to recruiter decisions beyond the resume and \emph{Resume Score}. Experiment~2 randomizes pipeline assignment at Stage~2: treatment-arm applicants are invited to the AI interview before any recruiter review, while control-arm applicants proceed directly to Stage~3. This allows us to analyze the effect of adopting an AI-assisted pipeline, inclusive of attrition at the AI stage and the resulting compositional changes in the pool recruiters evaluate.

The two designs also differ in the recruiter's working environment, which we view as a feature of the comparison. In Experiment~1, we closed the recruitment of candidates relatively fast and as a result each recruiter reviews a manageable number of candidates (the largest set presented to a recruiter is 266 candidates). This allows recruiters to evaluate every application in detail. In Experiment~2, we follow a standard procedure for hiring for the advertised position, which results in thousands of applicants reaching the recruiter's desk and heuristic filtering is unavoidable; treatment-arm recruiters can form a heuristic around the AI-interview signal by concentrating attention on candidates who completed the AI interview and who cleared the platform threshold for passing it; while control group recruiters might form heuristics based on the \emph{Resume Score}. Experiment~2 therefore varies both the recruiter's information and the decision rule it induces, while Experiment~1 holds the decision environment fixed and varies information alone. Together, the two designs produce a layered set of estimates: a cleaner information-channel estimate from Experiment~1, and a pipeline-level estimate from Experiment~2. In both experiments, the hiring managers interviewing candidates are blind to treatment assignment, do not observe the \emph{AI Interview Report}s, and are not aware of any details of the experiment; in particular, they do not know what was randomized and what is evaluated.

\subsection{Experiment 1: Randomizing the Recruiter's Information Set}
\label{sec:exp1-design}

To conduct Experiment~1, we advertised four job positions on LinkedIn: Frontend Engineer, Quality Assurance Engineer, Machine Learning Engineer, and Data Engineer. We closed recruitment when  4{,}323 applicants applied and 1,108 of them completed the AI interview. The experimental sample is those candidates who completed the AI interview; thus,  we observe the \emph{AI Interview Report} for all of them.

Within each job posting, subjects were randomized to a treatment or control group with equal probability. Randomization at Stage~3 ensures that the candidate pool reaching the recruiter is identical in expectation across arms, so any difference in downstream outcomes is attributable to the informational content of the AI report. Recruiters selected shortlists separately by role (Frontend, QA, Data Engineer, ML Engineer), with shortlist sizes set proportional to the number of completers in each role. A recruiter is reviewing only candidates from one experimental group. Thus, a control group recruiter will never see any candidates from the treatment group and vice versa. Finally, the hiring manager, who conducts an interview, receives an invitation to their calendar with a candidate resume, does not observe the \emph{AI Report}, and is blind to the candidate's experimental group assignment.

\subsection{Experiment 2: Randomizing Pipeline Assignment}
\label{sec:exp2-design}

For Experiment~2 we advertised a single Junior Frontend Engineer position requiring proficiency in React, JavaScript, and CSS. We followed the platform’s standard practice regarding the duration and scope of the posting. 34,493 unique candidates applied for the role and were randomized 70/30 in favor of the AI-assisted pipeline, with the unequal probabilities chosen ex ante to absorb the high anticipated attrition in the AI arm.\footnote{We received 36,838 applications. However, some candidates submitted more than one application, we drop all applications from those candidates. Keeping one application per candidate does not meaningfully alter treatment effect estimates.} 25{,}536 applicants were assigned to treatment and 8{,}957 to control. Randomization at Stage~2 means that treatment is defined as assignment to the AI-interview track, not completion of the AI interview itself.

Control-arm applicants are forwarded directly to a recruiter. Treatment-arm applicants are invited to complete the AI interview; 25.07~percent did so. The platform assigns each completer a binary pass label based on a fixed threshold: at least 'Mid-level' on each of the three required technical skills, a proctoring score of at least 70~percent, at least Mid-level soft skills, and at least two years of relevant experience, which 287 completers received. All applicants in this arm, regardless of the label, were forwarded to the recruiter, who observed the resume, \emph{Resume Score}, \emph{AI Interview Report}, and the pass label. We use the pass label in our analysis as a signal the recruiter could anchor on, not as a hard filter. The 19{,}133 invited applicants who did not complete the AI interview remain in the treatment arm were presented to the recruiter, and could in principle be shortlisted (even though in practice recruiters focus on candidates who completed and passed the AI interview). The non-completion is itself an outcome of treatment assignment and is analyzed in Section~\ref{sec:dropout}. Recruiters in both arms selected a fixed shortlist of 35 candidates, set by the number of hiring-manager interview slots available for the role.

Unlike Experiment~1, the candidate pool reaching the recruiter is not identical across arms. In particular, it depends on treatment-induced attrition. This is by design: Experiment~2 evaluates the full pipeline as a firm would deploy it. As a secondary outcome, we collect five-month LinkedIn employment data for a subsample.

\subsection{Variables and Data}
\label{sec:variables}

For each candidate we observe four classes of variables:

\textit{Treatment assignment.} The candidate's experimental arm (treatment or control) and the experiment to which they belong (Experiment 1 or Experiment 2).

\textit{Pre-treatment covariates, observed for all candidates.} The submitted resume, a keyword-based \emph{Resume Score} (0--100) generated by the platform for the job description, self-reported skills based on their resume, years of professional experience, highest level of education, position applied for, age, gender (inferred from name), country of residence, and application source (LinkedIn or hiring platform website). 

\textit{AI interview variables.} The \emph{AI Interview Report} described in Section~\ref{sec:platform}. In Experiment~1 the report is realized for every candidate in the experimental sample, and recruiters' \emph{access} to it varies by arm. In Experiment~2 the report is realized only for treatment-arm completers.

\textit{Outcomes.} AI-interview completion (where applicable), recruiter shortlisting at Stage~3, and the hiring-manager pass/fail decision at Stage~4. For Experiment~2 we additionally observe five-month LinkedIn employment outcomes for a subsample.

Balance across arms is reported in Appendix Tables ~\ref{tab:balance} and ~\ref{tab:summary_stats_treatment_balance}. In Experiment~2, standardized mean differences are below 0.1 for nearly all covariates. In Experiment~1, balance holds within each of the four roles for most covariates, with some role-by-covariate cells exceeding $|\text{SMD}| = 0.25$.
%\input{framework}
% =====================================================================
\section{Results}
\label{sec:results}
% =====================================================================

Recruiters who saw the \emph{AI Interview Report} shortlisted candidates who were substantially more likely to pass the final blind interview, both when the report's informational role was isolated (Experiment~1) and when the AI interview was deployed as a full pipeline filter (Experiment~2). Table~\ref{tab:main_ate} reports estimates from both experiments for two estimators: an unadjusted difference in means, and an augmented inverse-propensity-weighted (AIPW) estimator that adjusts for observed candidate characteristics. We report the results as percentage points increases. The AIPW estimator conditions on the candidate's resume score, position applied for, and age and gender, when available. We also report estimates under two codings of no-shows: first, we exclude them; second, we code them as failed interviews.\footnote{When a shortlisted candidate fails to attend the final interview, the platform's standard practice is to invite a replacement from further down the queue. Tracking outcomes for these replacement candidates was logistically complicated, so we exclude them from the analysis.}

\begin{table}[!htbp]
\centering
\caption{Effect of the \emph{AI Interview Report} and the AI-Assisted Pipeline on Final-Interview Pass Rate}
\label{tab:main_ate}
\begin{tabular}{l cccc}
\toprule\toprule
 & \multicolumn{2}{c}{\emph{No-shows excluded}} & \multicolumn{2}{c}{\emph{No-shows = 0}} \\
\cmidrule(lr){2-3}\cmidrule(lr){4-5}
 & Unadjusted & AIPW & Unadjusted & AIPW \\
 & (1) & (2) & (3) & (4) \\
\midrule
\multicolumn{5}{l}{\textbf{Panel A.\ Experiment~1:} information value} \\
\addlinespace[2pt]
$\widehat{\tau}^{\mathrm{info}}$
 & $0.204^{**}$  & $0.319^{***}$
 & $0.175^{**}$  & $0.287^{***}$ \\
 & $(0.097)$ & $(0.067)$ & $(0.085)$ & $(0.062)$ \\
Permutation $p$
 & $0.034$ & $0.0004$ & $0.034$ & $0.0006$ \\
\addlinespace[3pt]
Control mean         & $0.379$ & $0.379$ & $0.286$ & $0.286$ \\
Observations         & $108$   & $108$   & $140$   & $140$ \\
\midrule\midrule
\multicolumn{5}{l}{\textbf{Panel B.\ Experiment~2:} operational deployment} \\
\addlinespace[2pt]
  $\widehat{\tau}^{\mathrm{op}}$
   & $0.249^{**}$  & $0.303^{**}$
   & $0.200^{*}$   & $0.262^{**}$ \\
   & $(0.123)$ & $(0.123)$ & $(0.118)$ & $(0.117)$ \\
Permutation $p$
   & $0.083$ & $0.018$ & $0.149$ & $0.031$ \\
\addlinespace[3pt]
Control mean         & 0.364 & 0.364 & $0.343$ & $0.343$ \\
Observations         & $64$ & $64$ & $70$    & $70$    \\
\bottomrule\bottomrule
\end{tabular}

\vspace{6pt}
\begin{minipage}{\textwidth}
\footnotesize
\textit{Notes.} The outcome is an indicator for passing the final hiring manager interview. Columns~(1)--(2) drop candidates who were scheduled but did not attend the final interview; columns~(3)--(4) code those no-shows as failures. In Experiment~2, six finalists did not attend the final interview, so $N$ falls from 70 to 64 between codings. Unadjusted columns report stratified differences-in-means; weighted by position in Experiment~1. AIPW columns report augmented inverse-propensity-weighted estimates implemented with generalized random forests. Standard errors are in parentheses; for Panel~A they are clustered by position. Permutation $p$-values come from reshuffling treatment labels among finalists within position strata 5{,}000 times. Significance: $^{*}\,p<0.10$, $^{**}\,p<0.05$, $^{***}\,p<0.01$.

%\smallskip\noindent\textsuperscript{\textdagger}\,\textit{Temporary note.} The Panel~A AIPW columns currently report OLS estimates of $Y_i = \alpha + \tau\, G_i + r_i'\beta + \delta_p + \varepsilon_i$ (position fixed effects $\delta_p$ plus centered Resume Score with a missingness indicator), reported as ``AIPW'' here for symmetry with Panel~B pending replacement with a true AIPW estimator.
\end{minipage}
\end{table}

\subsection{Experiment 1: The Information Value of the Report}

Recruiters with access to the \emph{AI Interview Report} shortlisted candidates who passed the final blind interview at substantially higher rates than recruiters who saw resumes alone. The unadjusted estimate of the difference in pass rates is 20.4 percentage points (SE 9.7, $p_{\text{perm}} = 0.034$) when no-shows are excluded and 17.5~pp (SE 8.5, $p_{\text{perm}} = 0.034$) when they are coded as failures. Relative to control pass rates of 37.9 and 28.6~percent, these are 54 and 61~percent increases in final-interview pass rates. The adjusted estimates are larger: 31.9~pp (SE 6.7, $p_{\text{perm}} = 0.0004$) and 28.7~pp (SE 6.2, $p_{\text{perm}} = 0.0006$).

Adjusting for resume characteristics raises the estimate (from 17.5 to 28.7 pp). Relative to candidates shortlisted in the control arm, treatment-arm finalists have lower \emph{Resume Scores} by 5.55 points ($p<0.001$), but are 25 percentage points more likely to be rated at least Mid-level on all evaluated skills ($p<0.001$). The \emph{Resume Score} difference is large relative to the control-arm finalist mean of 92.09, while the skill-rating difference more than doubles the control-arm shortlisted candidates share of 0.23. This matches the design: control recruiters relied primarily on the resume-based signal, whereas treatment recruiters used the \emph{AI Interview Report} to identify candidates whose demonstrated skills exceeded what the resume alone would suggest.

The maximum \emph{Resume Score} is 96.5 in every one of the four roles, and only eight of the 1{,}108 completers (four of the finalists) reach it. The resume cannot separate the top of the pool, so recruiters shortlist almost entirely from below that ceiling. There the \emph{AI Interview Report} drives a wide skill gap: among finalists scoring below their role's maximum, treatment recruiters select candidates rated at least \emph{Mid-level} on all three technical skills 62~percent of the time, against 32~percent under resume screening ($p<0.001$), and at \emph{Resume Scores} about six points lower on average. By the final interview the gap closes, the blind hiring manager passes skilled candidates at about the same rate from each arm (near 54~percent), so the report does its work at the shortlisting margin that Experiment~1 randomizes, pulling a large group of demonstrably skilled candidates into contention that the \emph{Resume Score}, bunched near its ceiling, would not have separated.

\subsection{Experiment 2: Operational Deployment}

The AI-assisted pipeline also produced finalists who passed at higher rates. The unadjusted estimate of the pass rates difference is 24.9 percentage points (SE 12.3, $p_{\text{perm}} = 0.083$) excluding no-shows and 20.0~pp (SE 11.8, $p_{\text{perm}} = 0.149$) coding them as failures. Under the no-shows-as-failures coding, the control pass rate is 34.3~percent and the unadjusted contrast amounts to a 58~percent relative increase in finalist success rate. The AIPW estimates are larger: 30.3~pp (SE 12.3, $p_{\text{perm}} = 0.018$) and 26.2~pp (SE 11.7, $p_{\text{perm}} = 0.031$).\footnote{Expressed as effect sizes, the unadjusted estimates equal 0.42 (SE~0.20) and 0.39 (SE~0.19) standard deviations in Experiment~1 (columns~1 and~3) and 0.52 (SE~0.26) and 0.42 (SE~0.25) in Experiment~2. The unit is the control-group standard deviation of the pass indicator, $\sqrt{\bar{p}_c(1-\bar{p}_c)}$, computed within each no-show coding.}

Two important caveats shape the interpretation of the estimate. First, even though recruiters could shortlist any candidate from the treatment group, in practice they do not consider candidates who dropped out from the AI interview. Section~\ref{sec:dropout} documents pool composition and unobservable-selection concerns. Second, recruiters selecting from the treatment group had a larger consideration pool than recruiters in the control group; thus, the pipeline-level contrast could partly reflect an order statistic. In \ref{sec:null_pool}, we consider two adjustments: first, we simulate candidate pools with quality drawn from a normal distribution; second, from an empirical distribution of \emph{Resume Score}. Under the former scenario, such mechanical sample size effect creates a 4.5 p.p. difference in the second stage pass rates. In the latter scenario, there is no difference in the second stage, because there are more than 35 candidates with a perfect \emph{Resume Score} in each experiment's arm.

% =====================================================================
\section{Mechanisms}
\label{sec:mechanisms}
% =====================================================================

In this section, we present two mechanisms associated with the observed differences. The first channel is verification: the AI interview disciplines skill claims that resumes routinely overstate. The second is participation: the interview imposes a cost on applicants and so reshapes who remains in the pool.

\subsection{Mechanism 1: Resume vs.\ AI-Vetted Skill Verification}
\label{sec:lying}

Software-engineering resumes routinely list a long catalog of technologies. In our experimental sample candidates listed 23 skills on average. Prior work documents that more than half of applications contain deliberate skill inflation \citep{henle2019assessing}, and the Dunning-Kruger bias implies that candidates often overstate their own proficiency \citep{kruger1999unskilled}. A natural question is whether the \emph{AI Interview Report}'s predictive content comes in part from detecting these inflated claims.

The job listing we posted required advanced proficiency in React, JavaScript, and CSS, and these three skills were evaluated during the AI interview. In this section, we consider cases in which the candidate explicitly mentions proficiency in the skill, while receiving the \emph{Not Familiar} grade from the AI interview, which is the lowest grade and reflects cases in which the candidate did not respond correctly to any questions measuring the skill.

\paragraph{Evidence from the experiment.} For each treatment-arm completer in Experiment~2 we flag a candidate as \emph{misreporting} a skill if a required technology - React, JavaScript, or CSS -appears in their resume and the AI scores the candidate's proficiency in it as \emph{Not Familiar}. Table~\ref{tab:skill_deficiency} reports the rates. Roughly 21~percent of completers fail this minimal screen on at least one of the three required skills (SE 0.5 pp.); 12~percent on two; 7.5~percent on all three. A resume that lists React, JavaScript, and CSS therefore conveys, with non-negligible probability, that the candidate cannot demonstrate any of them.

\paragraph{Replication on a historical audit.} A natural concern is that this rate is specific to Experiment~2's three skills or to the treatment arm's particular pool. We replicate the exercise on a separate, larger sample drawn from micro1's historical platform data: we match AI skill ratings to resume strings for 720 candidates across multiple roles and skill sets, and flag a misreport whenever a required technology appears on the resume but the AI rates it as \emph{Not Familiar}. The audit recovers a similar rate: 21.6 percent on at least one skill (SE 1.5 pp.), 10.4 percent on two, 5.2 percent on all three, suggesting that the verification value of the \emph{AI Interview Report} is not an artifact of Experiment~2's particular role.

\begin{table}[!htbp]
\centering
\caption{Proportion of Candidates with Skill Misreporting}
\label{tab:skill_deficiency}
\begin{threeparttable}
\begin{tabular}{lcc|cc}
\toprule\toprule
& \multicolumn{2}{c}{Experiment 2 completers} & \multicolumn{2}{c}{Historical audit} \\
\cmidrule(lr){2-3}\cmidrule(lr){4-5}
Number of misreported skills & Share & S.E. & Share & S.E. \\
\midrule
One or more & $0.213$ & $(0.005)$ & $0.216$ & $(0.015)$ \\
Two or more & $0.124$ & $(0.004)$ & $0.104$ & $(0.012)$ \\
Three       & $0.075$ & $(0.003)$ & $0.052$ & $(0.010)$ \\
\bottomrule\bottomrule
\end{tabular}
\begin{tablenotes}\footnotesize
\item \textit{Notes:} A misreport is flagged when a required technology appears on the resume but the \emph{AI Interview Report} rates the candidate as \emph{Not Familiar}. Columns~1--2 use Experiment~2 treatment-arm completers and the three required skills (React, JavaScript, CSS). Columns~3--4 use a random audit of 720 historical micro1 candidates across multiple roles and skill sets. Standard errors in parentheses.
\end{tablenotes}
\end{threeparttable}
\end{table}

\begin{figure}[!htbp]
\centering
\caption{Heterogeneity in Misreported Skills Across Candidate Subgroups}
\label{fig:skill_hte}
\includegraphics[width=0.95\textwidth]{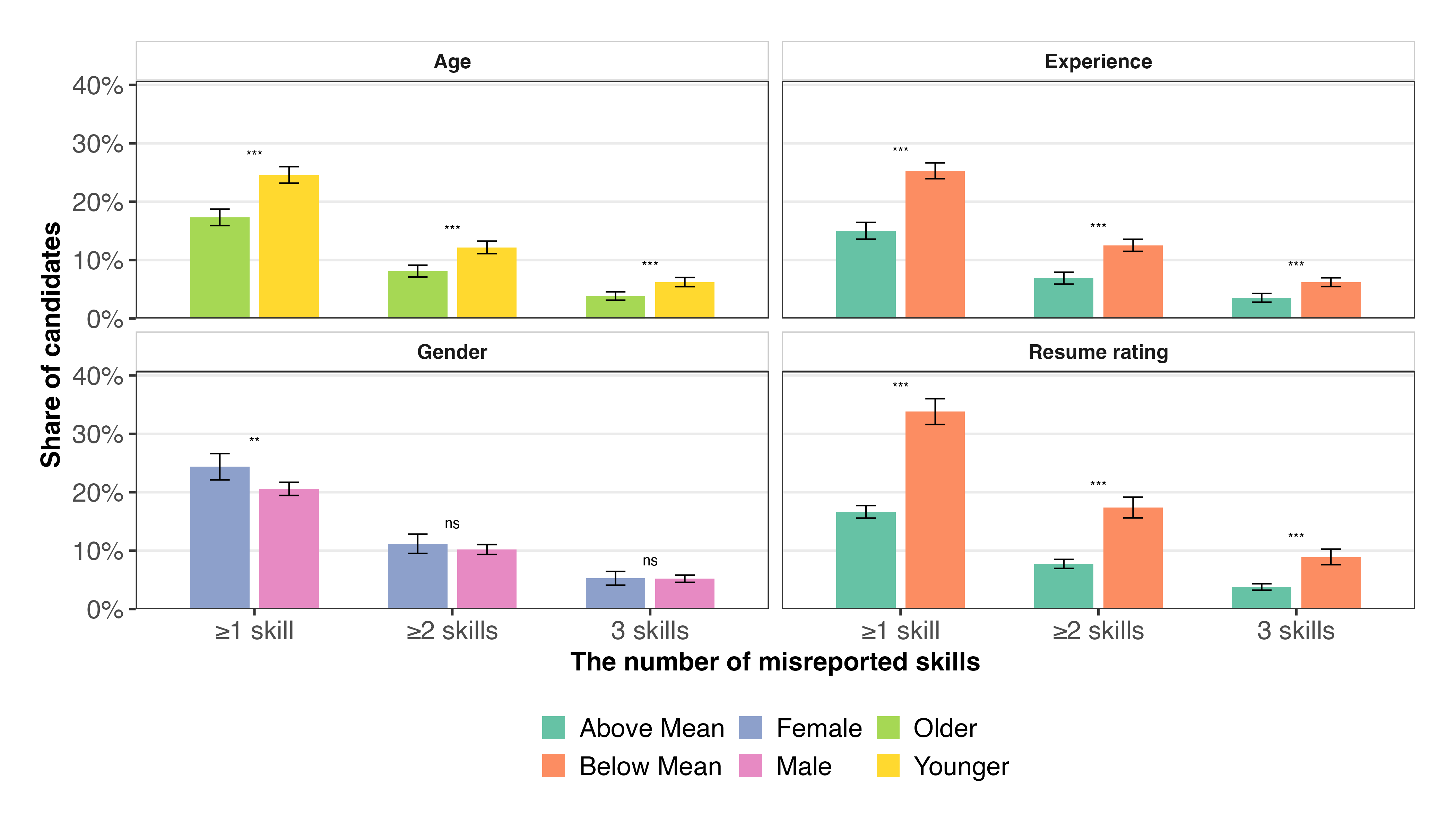}
\vspace{0.3em}
\begin{minipage}{0.92\textwidth}
\footnotesize\textit{Notes:} Bars show the share of candidates who misrepresented at least one, at least two, or all three required technologies (React, JavaScript, CSS); whiskers denote 95~percent confidence intervals. Age and experience are split at sample means; resume scores at the mean automated score; gender is inferred from names.
\end{minipage}
\end{figure}

\paragraph{Heterogeneity in misreporting.} Misreporting is concentrated in the candidate segments where verification is most difficult under text-only screening. Figure~\ref{fig:skill_hte} decomposes the rate by age, experience, resume score, and gender. Younger and less-experienced candidates and those with lower \emph{Resume Scores} misreport at higher rates; gender differences are small and statistically insignificant once we look at two or more skills. The implication is that the \emph{AI Interview Report} adds disproportionate verification value on the candidate segments that pose the greatest verification challenges based on resume.

\paragraph{External validation.} We validate the AI skill ratings using an independent external validator who was unaffiliated with micro1 and blind to the AI scores. The validator, a senior full-stack developer hired through Upwork, reviewed a stratified random sample of 60 AI interview transcripts from Experiment 2 and independently coded whether each candidate demonstrated React, JavaScript, and CSS skills. Across 180 transcript-skill ratings, the validator agreed with the AI assessment in 96.1\% of cases (Wilson 95\% CI [92.2\%, 98.1\%]). In the seven disagreements, the validator rated the candidate as ``Not Familiar," suggesting that our estimates of skill misrepresentation are, if anything, conservative. \ref{audit_info} provides additional details on the sampling procedure and validation results.

\subsection{Mechanism 2: Selective Participation}
\label{sec:dropout}
% maybe we should phrase this as "people who are actively looking for a job are more likely to complete the interview," because this is mapping better to our LinkedIn results, right? You can say, "Oh, they're looking for other jobs too." One thing is that maybe they're better candidates, but also maybe that means that they are actively pursuing a new job 
Completing the AI interview takes 30 to 40 minutes, as a consequence only 25\% of candidates invited for the interview in Experiment 2 completed and the resulting dropout is not random. We characterize it on three margins: (i)~on what observable characteristics dropout selects, (ii)~what happens to dropouts in the labor market afterwards, and (iii)~how the resulting change in pool composition shifts who gets advanced relative to a resume-only baseline.

\begin{figure}[!htbp]
\centering
\caption{Distribution of Resume Scores}
\label{fig:selection_attrition}
\includegraphics[scale=0.27]{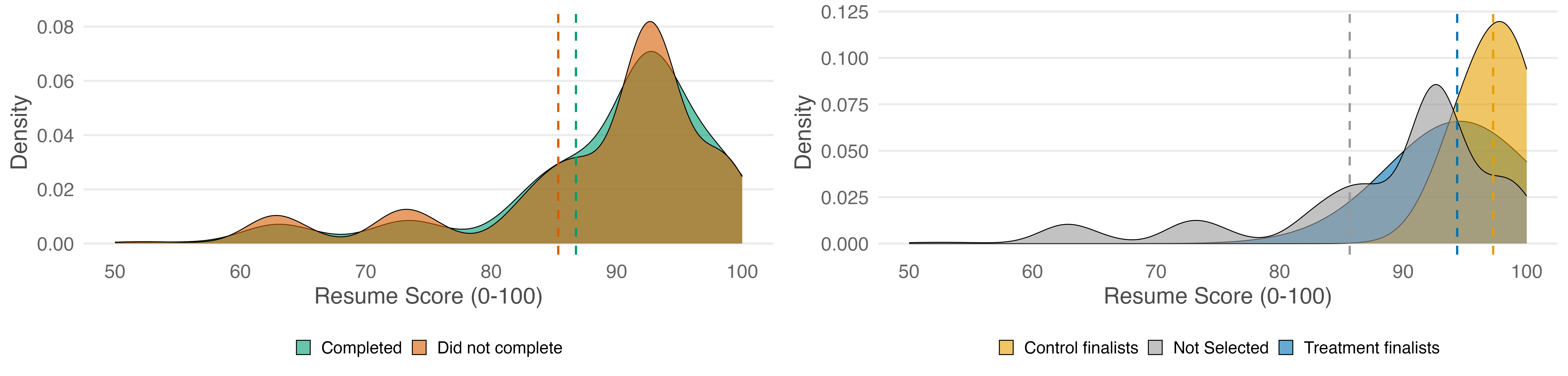}
\vspace{0.3em}
\begin{minipage}{\textwidth}
\footnotesize\textit{Notes:} Left panel: density of \emph{Resume Score} for treatment-arm candidates who completed the AI interview against those who were invited but did not. Right panel: \emph{Resume Score} densities for control-arm finalists, treatment-arm finalists, and the pool of candidates not selected for the human interview.
\end{minipage}
\end{figure}

\paragraph{(i) On what dropout selects.} Figure~\ref{fig:selection_attrition} shows the distribution of \emph{Resume Scores} across key subsets. The left panel compares treatment-arm completers against treatment-arm non-completers: the two distributions are similar, with completers having only modestly higher scores on average (difference is 1.41 points, SE 0.243). The right panel compares \emph{Resume Scores} of candidates selected in the treatment group, selected in the control group, and not selected. Recruiter-selected candidates in both arms have substantially higher \emph{Resume Scores} than non-selected candidates (difference is 10.7 points, SE 0.407), but the recruiter-selected treatment-arm candidates have somewhat lower \emph{Resume Scores} than recruiter-selected control-arm candidates (difference is -2.92 points, SE 0.762). The last comparison is consistent with treatment recruiters substituting AI-assessed skill for resume signal.

Table~\ref{tab:dropped_out_estimates} regresses an indicator for non-completion on observed characteristics. Years of experience, education, gender, and \emph{Resume Score} all predict dropout. Each additional year of experience raises the probability of non-completion by 0.3--0.4~pp; holding a bachelor's degree lowers it by roughly 2.7~pp relative to other education levels; men drop out about 1.3--1.5~pp less than women ($p < 0.05$); and one additional \emph{Resume Score} point is associated with 0.1 percentage-point lower dropout probability ($p < 0.001$). When we instead use the candidate's predicted probability of passing the interview---trained on the subsample that did complete---the predicted-pass coefficient is small and insignificant once \emph{Resume Score} and other observables are included (column~4). Dropout therefore selects on resume quality and demographics; it does not appear to select on a candidate's \emph{predicted} ability to pass the interview itself, conditional on those observables.

\begin{table}[!htbp]
\centering
\caption{Estimates of the Probability of Dropping Out}
\label{tab:dropped_out_estimates}
\resizebox{\textwidth}{!}{%
\begin{tabular}{@{\extracolsep{5pt}}lcccc@{}}
\toprule\toprule
 & \multicolumn{4}{c}{\textit{Dependent variable: Dropped out}}\\
\cline{2-5}
 & OLS & OLS + Resume Score & OLS w/ Prob.\ Pass & OLS w/ Prob.\ Pass + Cov. \\
\midrule\\[-1.2ex]
Probability to Pass & & & $-0.210^{*}\,(0.114)$ & $0.090\,(0.138)$ \\
Years of Experience & $0.003^{***}\,(0.001)$ & $0.004^{***}\,(0.001)$ & & $0.003^{***}\,(0.001)$ \\
Age                 & $0.0003\,(0.001)$       & $0.001\,(0.001)$        & & $0.001\,(0.001)$ \\
High School         & $-0.005\,(0.035)$       & $-0.009\,(0.035)$       & & $-0.010\,(0.035)$ \\
Bachelor            & $-0.027^{***}\,(0.006)$ & $-0.026^{***}\,(0.006)$ & & $-0.026^{***}\,(0.006)$ \\
Master's            & $-0.007\,(0.009)$       & $-0.004\,(0.009)$       & & $-0.007\,(0.009)$ \\
Male                & $-0.015^{**}\,(0.007)$  & $-0.013^{**}\,(0.007)$  & & $-0.013^{*}\,(0.007)$ \\
Resume Score        &                          & $-0.001^{***}\,(0.0002)$ & & $-0.001^{***}\,(0.0002)$ \\
Constant            & $0.757^{***}\,(0.023)$  & $0.839^{***}\,(0.027)$  & $0.759^{***}\,(0.006)$ & $0.834^{***}\,(0.028)$ \\
\midrule\\[-1.2ex]
Observations & 24{,}870 & 24{,}870 & 25{,}504 & 24{,}870 \\
$R^2$        & 0.002 & 0.004 & 0.0001 & 0.004 \\
\bottomrule\bottomrule
\end{tabular}}
\vspace{0.3em}
\begin{minipage}{\textwidth}
\footnotesize\textit{Notes:} OLS estimates of the probability of dropping out of the AI interview on observed characteristics. The predicted probability of passing the AI interview is estimated by gradient boosting on the subsample of completers, following \citet{athey2025machine}. Standard errors in parentheses. Significance: $^{*}p<0.10$; $^{**}p<0.05$; $^{***}p<0.01$.
\end{minipage}
\end{table}

\paragraph{(ii) What happens to dropouts afterwards.}
  
Five months after treatment assignment, we collected LinkedIn profiles for 3{,}641 applicants, including all shortlisted candidates and those with the highest \emph{Resume Score}s, and recorded whether each candidate listed a new position starting after the experiment began.\footnote{We consider any new job with a start date after the experimental group assignment. This includes new jobs at firms in which applicants had previous employment.} 

\begin{figure}[!htbp]
\centering
\caption{New Jobs Reported on LinkedIn by Group}
\label{fig:linkedin_groups}
\includegraphics[scale=0.5]{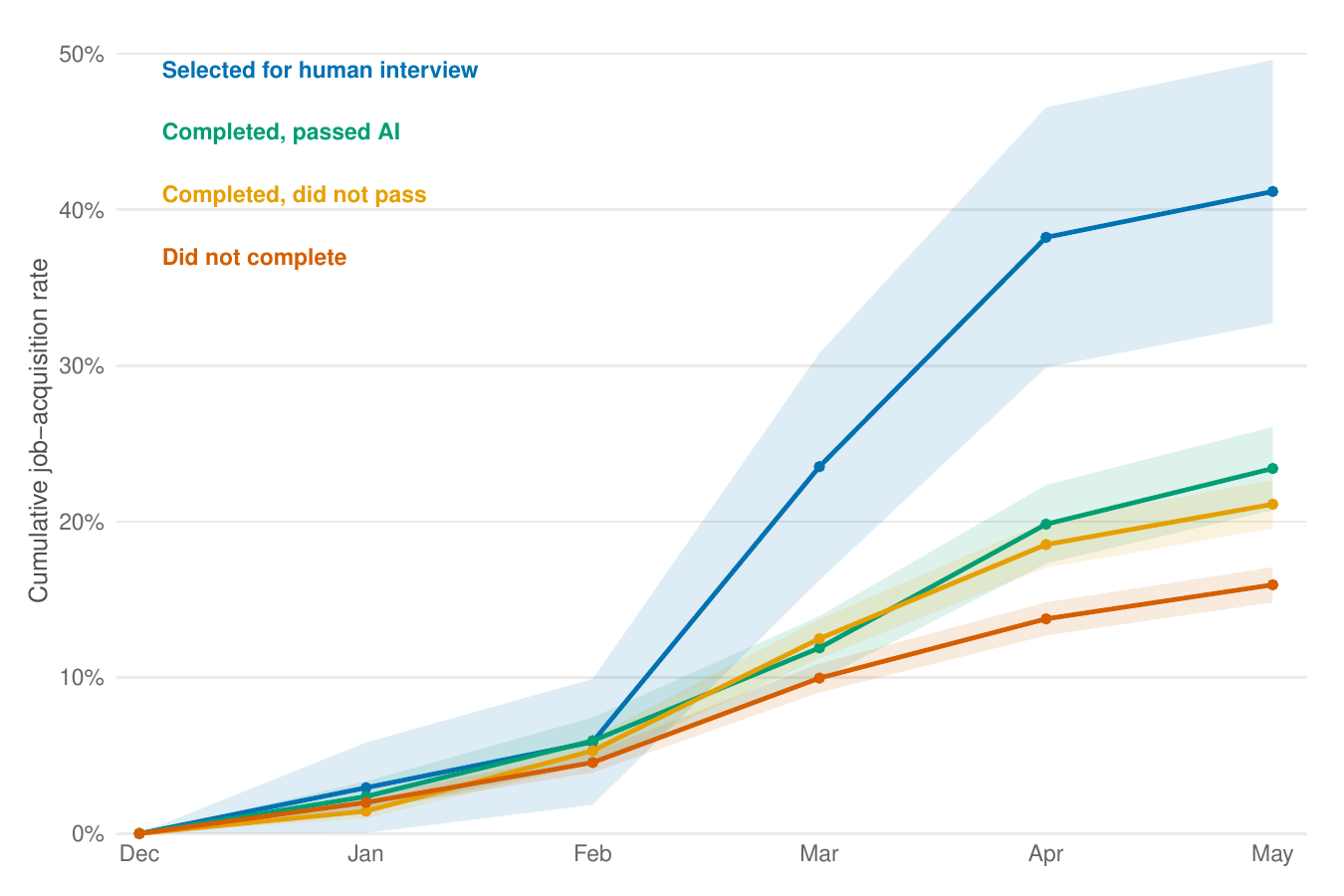}
\vspace{0.3em}
\begin{minipage}{\textwidth}
\footnotesize\textit{Notes:} Shares of candidates who reported a new job on LinkedIn across four subsets of the treatment arm: did not complete the AI interview (orange), completed but did not pass (yellow), completed and passed (green, includes shortlisted candidates), and completed, passed, and were selected for the human interview (blue). Shaded areas denote standard errors.
\end{minipage}
\end{figure}

Figure~\ref{fig:linkedin_groups} presents subsequent employment rates for shortlisted candidates, treatment-arm non-completers, treatment-arm completers, and the subset of completers who failed the AI interview. The shortlisted candidates had a 41 percent chance of reporting a new job; a rate that is substantially higher than any other group. Non-completers reported new jobs substantially less often than completers: 16.0 percent versus 22.4 percent, a 6.5-percentage-point gap (SE 1.7). The gap persists even among candidates who did not pass the AI interview. Treatment-arm completers who failed the AI interview reported new jobs at a higher rate than non-completers: 21.1 percent versus 16.0 percent, a 5.2-percentage-point gap (SE 1.9).

The last comparison points to selection: both groups failed to advance through the AI-assisted pipeline, but they differ in whether they completed the 30--40 minute assessment. The higher subsequent employment rate among failed completers suggests that completion selects on unobserved labor-market activity, motivation, or job-search intensity, beyond the observable resume and demographic margins documented above.

%We also compare \emph{New Job} rates across the experimental arms. Among the 70 candidates advanced to the human interview, 39~percent of treatment-arm finalists reported a new job five months later, against 22~percent of control-arm finalists---a gap of 17~percentage points (SE~8.2). Widening the comparison to all treatment-arm candidates who passed the AI interview against all control-arm candidates who received the top \emph{Resume Score} preserves the direction of the effect at a 5.9-pp gap (SE~2.9).

\paragraph{(iii) How completion changes the advanced-candidate pool.} Dropout is correlated with observed characteristics; thus, even though the experimental samples were initially balanced, the candidate pools effectively considered in each arm differed (as documented above). Furthermore, recruiters in each treatment arm had access to different signals about candidates. The two selections compound to differences in candidates shortlisted (or likely to get shortlisted across the two arms). 

To make this concrete, we rank treatment-arm candidates by \emph{AI Score} and by \emph{Resume Score} and group them into terciles by the change in rank. Candidates in the top tercile (``benefits from AI'') have 2.7 years of experience on average against 5.6 in the bottom tercile, are younger (24.2 vs.\ 27.0), have lower \emph{Resume Scores} (72.0 vs.\ 97.6), and are less likely to hold a master's degree. \ref{app:rank_hte} reports the full tercile decomposition. 

\section{Where Do the AI Scores Add Predictive Content?}
\label{sec:auc}

Experiments 1 and 2 cover four positions—Frontend Engineer, QA Engineer, Data Engineer, and Machine Learning Engineer—closely related junior technical roles. Section \ref{sec:mechanisms} documented heterogeneity within this pool in which candidates benefit most from the \emph{AI Interview Report}. Here we ask a separate but related question: for which position types the AI interview adds the most value. To answer it, we draw on historical micro1 data covering a wider range of roles in which the AI interview was deployed, and we examine for which role types the \emph{AI Interview Report} is most useful in explaining the outcome of the final hiring-manager interview. The exercise is observational and conditional on advancing through the recruiter screen.

\paragraph{Setup.} We use a historical sample of candidates who completed both an AI interview and a final human interview. This sample encompasses 4,372 candidates who applied for 183 different roles. Thus, this is a sample of candidates whose AI interview result was high enough for a recruiter to advance them to the next stage.

We train two gradient-boosted classifiers on a stratified half of the sample and evaluate on the held-out half. The \emph{baseline} model uses \emph{Resume Score}, seniority, functional category, gender, country, and application date. The \emph{AI-augmented} model adds the outcome of the AI interview. Both models predict a pass or fail outcome from the hiring-manager interview; the hiring manager typically does not access the \emph{AI Interview Report} (it's a recruiter facing tool); however, we cannot rule out that in some cases a recruiter would pass on some information from the AI interview itself to the hiring manager. Predictive accuracy of both models is summarized by AUC.

\paragraph{Results.} Table~\ref{tab:auc} reports the results. In the top panel, we validate this approach using the samples from the two experiments. In Experiment 1 we have the \emph{AI Interview Report}s for both treatment arms, in Experiment 2 only for the treatment group; thus, the second sample is only 35 observations. We find that in both cases the AUC is higher for the model trained with the \emph{AI Interview Report}. The difference is statistically significant for the Experiment 1 data.

In the bottom panel, we present the results from the historical data. First, we show that the AUC is substantially higher for the model with the \emph{AI Interview Report}. Second, we group position into junior roles and non-junior ones.\footnote{We assign each candidate a seniority level from the platform's job-role title. The title is matched against two manually curated lists: one of senior-level titles (those prefixed ``Senior'' or ``Sr,'' ``Founding'' roles, and managerial or lead positions) and one of junior-level titles (individual-contributor engineering roles and the platform's skill-named screening interviews). A candidate is coded \emph{Junior} if the title appears on the junior list and \emph{Non-junior} otherwise; the Non-junior group therefore pools titles on the senior list with titles that match neither list. Matching is exact at the title-string level, so a title absent from both lists is treated as Non-junior. In the held-out half of the historical sample this yields 497 junior and 1{,}686 non-junior candidates.} Among the 497 junior candidates, AUC rises from 0.616 to 0.794 ($\Delta = 0.178$). In contrast, the non-junior roles AUC increases by 0.081 for the non-junior roles. The difference in the AUC gains is 0.098 (SE 0.045).

\begin{table}
\centering
\caption{Predictive Lift from AI Interview Report}
\label{tab:auc}
\begin{threeparttable}
\begin{tabular}{l ccccc}
\toprule\toprule
 & & & \multicolumn{2}{c}{AUC} & \\
\cmidrule(lr){4-5}
Subgroup & $N$ & Pass rate & Baseline & With AI & $\Delta$ AUC \\
\midrule
\multicolumn{6}{l}{\emph{Experimental samples}} \\
Experiment~1 (4-role RCT)     & $140$     & $0.364$ & $0.644$ & $0.734$ & $0.090^{*}$  \\
Experiment~2 (single role)    & $35$      & $0.543$ & $0.559$ & $0.663$ & $0.104$      \\
\midrule
\multicolumn{6}{l}{\emph{Historical platform sample}} \\
Overall                       & $2{,}183$ & $0.261$ & $0.796$ & $0.862$ & $0.066^{***}$ \\
Juniors                       & $497$     & $0.101$ & $0.616$ & $0.794$ & $0.178^{***}$ \\
Non-juniors                   & $1{,}686$ & $0.308$ & $0.776$ & $0.857$ & $0.081^{***}$ \\
\bottomrule\bottomrule
\end{tabular}
\begin{tablenotes}\footnotesize
\item \textit{Notes:} Out-of-sample AUC for predicting final-interview pass. Baseline features for the historical sample are Resume Score, seniority, functional category, gender, country, and application date; the AI-augmented model adds the counts of skills rated Senior, Mid-level, and Junior. The historical models are gradient-boosted classifiers trained on a stratified 50~percent split and evaluated on the held-out 50~percent. The Experiment~1 row uses the same feature set in a 5-fold cross-validated GBM on the 4-role RCT finalist sample. The Experiment~2 row is restricted to the treatment-arm finalists (the only subjects for whom AI-interview ratings exist by design); features are a compact logistic specification (gender, resume score, years of experience, AI skill counts) fit with 5-fold cross-validated out-of-fold predictions. Significance from DeLong's test: $^{*}\,p<0.05$, $^{**}\,p<0.01$, $^{***}\,p<0.001$. 
\end{tablenotes}
\end{threeparttable}
\end{table}

  \begin{figure}[!htbp]
  \centering
  \caption{ROC Curves: Junior vs.\ Non-Junior (Held-Out Test Set)}
  \label{fig:auc_seniority}
  \includegraphics[width=0.4\textwidth]{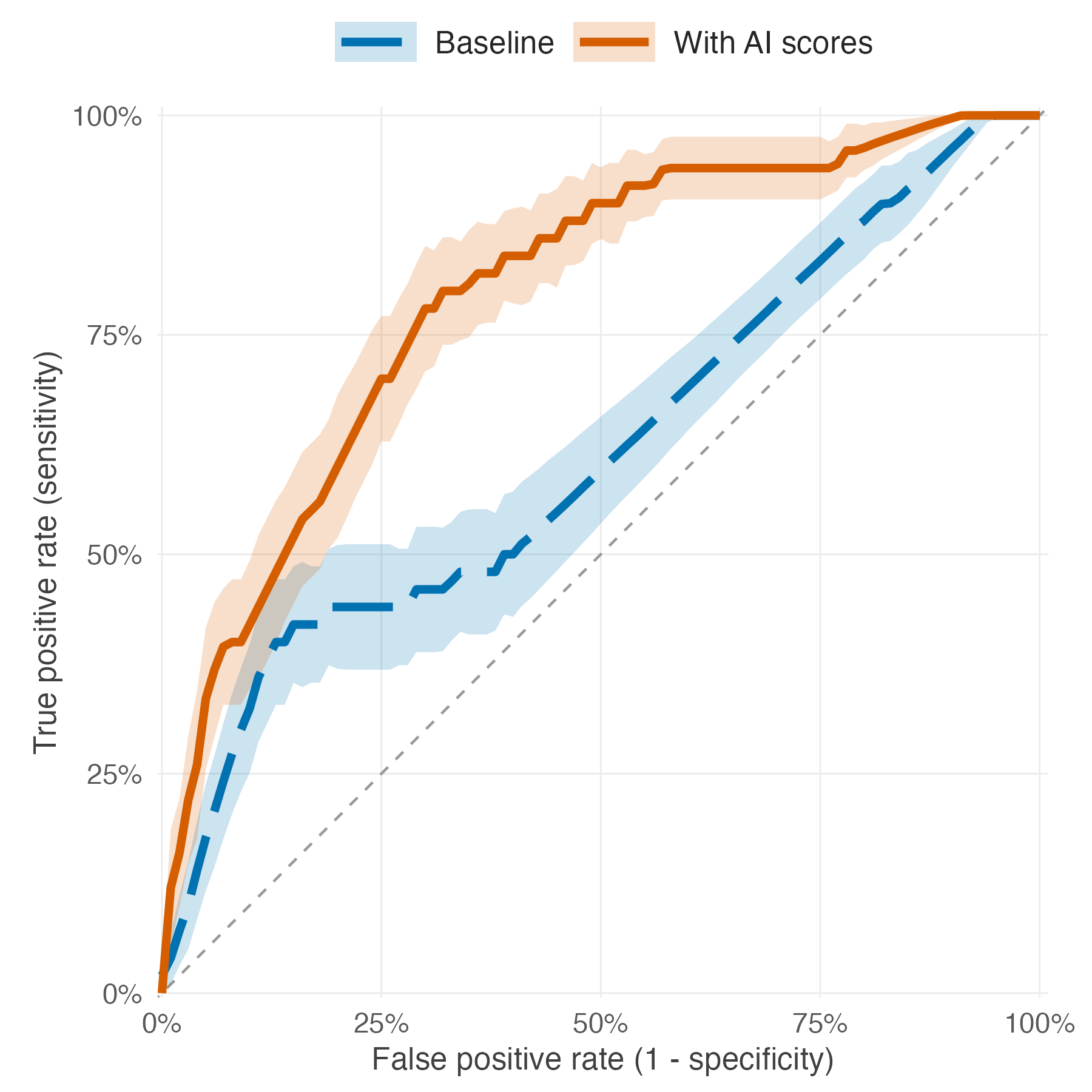}\hfill
  \includegraphics[width=0.4\textwidth]{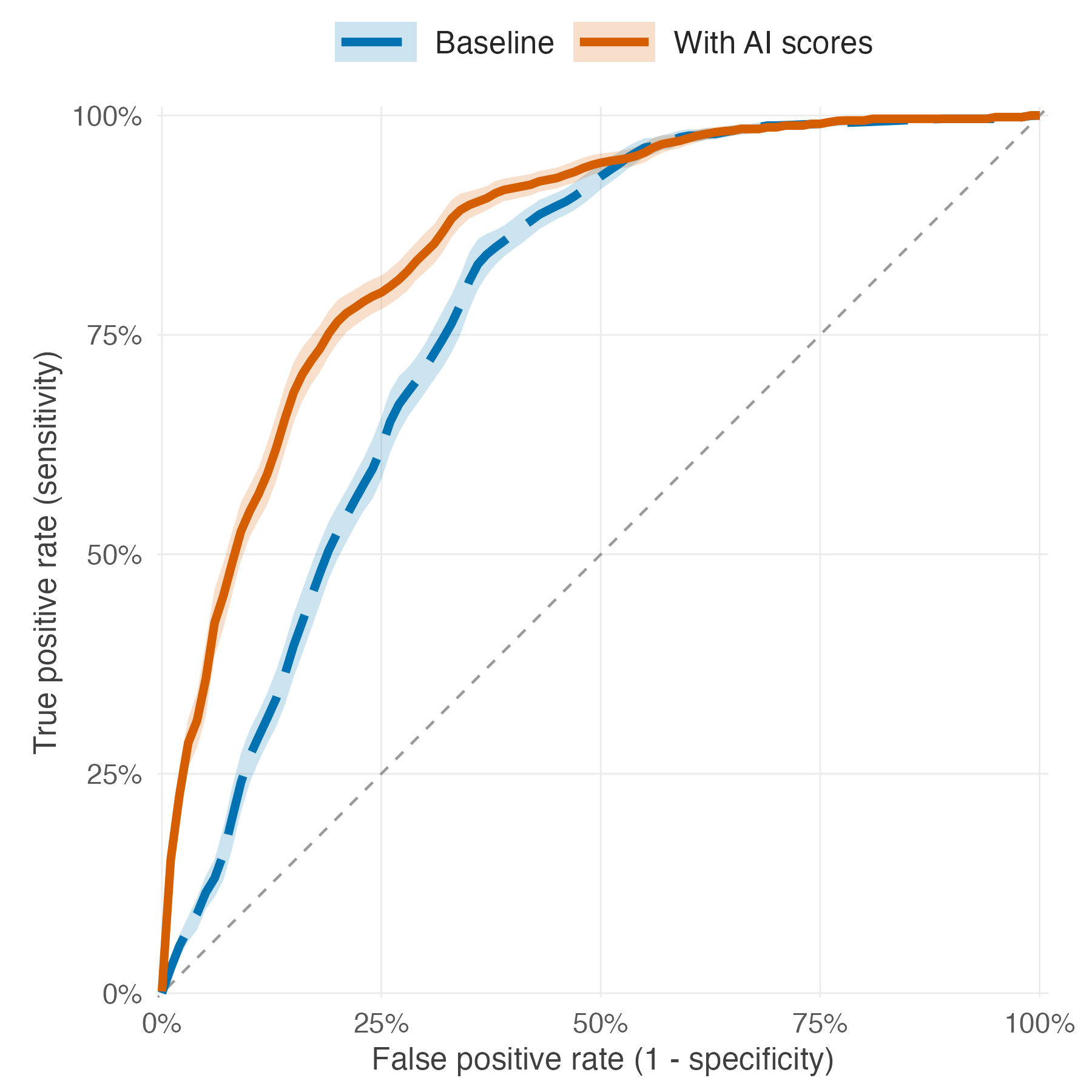}
  \parbox{0.95\textwidth}{\footnotesize\textit{Notes:} Out-of-sample ROC curves for the held-out 50\% of the historical platform sample. Left panel: junior candidates. Right panel: non-junior candidates. Shaded bands are 95\% pointwise confidence intervals
  computed via the DeLong method.}
  \end{figure}

Figure~\ref{fig:auc_seniority} plots out-of-sample ROC curves separately for juniors (left panel) and non-juniors (right panel). For juniors, the AI-augmented curve sits above the baseline at every threshold, with the largest separation at false-positive rates between 0.1 and 0.4. The low false-positive range is where a recruiter building a top-of-pipeline shortlist operates. For non-juniors, the two curves are closer together, though, their confidence intervals still do not overlap in the 0 -- 0.5 range. Resumes alone carry strong signals when candidates have long track records and conventional credentials, and weak signals when they do not. A 30--40~minute structured interview adds the most predictive content where the resume adds the least.
  
\paragraph{Caveats.} Two limits bound this exercise. First, AUC is computed on candidates who advanced past recruiter review; it does not estimate predictive performance on the full applicant pool. Second, even though the standard practice is that the \emph{AI Interview Report} is the recruiter facing tool; in some cases, the recruiter might pass the \emph{AI Interview Report} or some summary of it to the hiring manager; if so, the hiring manager might use the \emph{AI Interview Report} itself rather than their own judgment when deciding whether the candidate passed or failed. However, combined with the experimental evidence, the AUC analysis pins down where in the job and candidate distribution the report contributes: the experiments and the AUC analysis agree that AI-assisted interviewing earns its largest returns on junior, technical roles, which is the segment our experiments were run on.

% =====================================================================
\section{General Discussion}
\label{sec:discussion}
% =====================================================================
Our two field experiments show that the introduction of an AI-led structured interview into the recruitment pipeline raises the rate at which shortlisted candidates pass the final human interview. Experiment 1 holds the candidate pool fixed and varies only what the recruiter sees, isolating the informational value of the \emph{AI Interview Report}; Experiment 2 embeds the same interview in a live pipeline and varies pipeline assignment at the point of application. The two estimates are similar in magnitude. Behind these reduced-form effects, we identify two mechanisms: the AI interview disciplines skill claims that resumes routinely overstate, and it imposes a participation cost that reshapes who remains in the pool. Drawing on historical platform data, we further show that the predictive content of the \emph{AI Interview Report} concentrates among junior positions, precisely where candidates' conventional signals carry the least information.

\subsection{Practical Implications for Organizations}

This evidence points to several actionable take-aways for recruitment practice. Structured AI interviews add the most predictive content where resumes add the least. Adding \emph{AI Interview Report} skill ratings to a baseline of conventional resume features raises out-of-sample AUC by 0.18 for junior candidates as compared to the lift of 0.081 for non-junior candidates. A senior candidate's resume may already encode much of what a 30 to 40 minute structured interview can elicit; a junior candidate's resume often does not provide sufficient and trustworthy information. Firms should therefore target AI interviewing at top-of-funnel, early-career, and technical roles, rather than deploying it uniformly across the hiring pipeline.

The AI interviews also help convert resume claims into verifiable information. In Experiment 2, 21 percent of treatment-arm completers list at least one required skill that the AI rates as \emph{Not familiar}; 12 percent list two skills, and 7.5 percent misreport all three. We find a nearly identical rate in an audit of 720 historical candidates. The classification also appears reliable: in a random subset of 60 interview transcripts, an external validator agreed with the AI's skill classifications in 96.1 percent of cases, and all disagreements were cases in which AI was more lenient. These results suggest that the AI interview is an effective tool for verifying skills claims. By measuring skills before candidates reach the final stages of the funnel, the firm can avoid shortlisting candidates whose resumes contain inflated skills claims.

Requiring the AI interview induced substantial selective participation: only 25.07 percent of invited applicants completed the 30–40 minute assessment. Completion appears to capture information beyond assessed skill. Observable characteristics explain less than one percent of variation in noncompletion, and, conditional on these characteristics, predicted interview success is not significantly associated with completion. On the other hand, candidates who completed but failed the assessment were more likely to report a new job five months later than noncompleters (21.1 versus 16.0 percent), suggesting that there is an unobserved dimension that completion selects on. Such unobserved dimension could be reflecting job-search motivation, or willingness to incur application costs, although our data do not distinguish among these explanations. The AI interview therefore generates two distinct sources of information: whether a candidate completes it and what the resulting report reveals about demonstrated skill. 

Despite the low participation, the adoption of the AI interview reallocates the burden of an initial screen from recruiters to applicants. In Experiment 2, roughly 6,400 treatment-group applicants spent about 40 minutes each on the structured interview - an aggregate of 4,170 candidate-hours - while control applicants spent only a few minutes uploading a resume. The firm-side savings move in the opposite direction. Under a standard screening-round routine benchmark in which recruiters examine 200 applications at a time and stop once 35 hiring manager interview spots are filled, recruiter workload fell from about 160 to 82 hours because the AI screen raised the first-round pass rate from 34 to 54 percent. At a fully-loaded wage of \$55 per hour, the implied \$4{,}300 in labor savings exceeds typical platform subscription fees. The exchange rate, however, is steep: each recruiter hour saved corresponds to roughly 53 candidate hours. Taken together, managerial decisions to adopt the AI interview should therefore consider not only recruiter productivity, but also which groups of job seekers are screened in or out by the added procedural hurdle.

\subsection{Limitations and Scope}

Four limitations bound the reading of our results. The first is scope: both experiments cover junior technical roles, and the predictive analysis suggests that this is exactly the segment where AI assessments add the most. We therefore read our experimental magnitudes as the upper end of what an AI interview can deliver, with smaller gains expected on senior, leadership, or domain-specific roles.

The second concerns outcomes. We observe interview pass rates and a five-month LinkedIn employment proxy, but not post-hire productivity, retention, or job performance. Whether selection gains translate into matched-quality gains is the question that most directly bears on welfare, and longitudinal studies tracking on-the-job outcomes would provide direct evidence on this margin. Furthermore, although recruiters do not manage selected candidates, greater human involvement in identifying and advancing candidates could also affect perceptions of process ownership, confidence in the eventual hire, or coordination between recruiters and hiring managers. Whether such effects influence subsequent hiring decisions or employee outcomes remains an open question.

The third concerns identification within the information channel. Our Experiment 1 estimate does not separate the report's content from recruiters' response to it; an \emph{AI Interview Report} that is heavily weighted and one that is largely ignored produce different shortlists. The predictive analysis in Section \ref{sec:auc} partially addresses this by holding recruiter behavior fixed, but only on the recruiter-advanced subpopulation.

Finally, our results are based on small samples of finalists who participated in the interview with the hiring manager. Although the point estimates are substantial and statistically significant; we cannot rule out estimates that are much smaller. Furthermore, as a result, we cannot robustly compare the estimates from Experiment 1 and Experiment 2.

\subsection{Conclusion}
Across the two experiments, finalist sets generated with AI-interview information had final-interview pass rates approximately 20 percentage points higher than finalist sets generated from resume-based screening. The lift is largest where resumes are weakest: adding AI ratings raises out-of-sample AUC by 0.18 for junior candidates, against a lift of 0.08 for non-junior candidates. The main implication is operational: Our evidence suggests that AI interview reports may be especially informative in early-career technical hiring, where resumes provide relatively weak signals. Whether similar gains extend to senior or nontechnical roles remains an open question.

\clearpage
\bibliographystyle{informs2014}
\bibliography{refs}

\clearpage
\renewcommand{\thesection}{Appendix \Alph{section}}
\setcounter{section}{0}

\section{AI Interview}
\begin{figure}
    \centering
    \caption{AI Recruiter Platform}
    \label{fig:screen_shots}
    \includegraphics[width=0.83\linewidth]{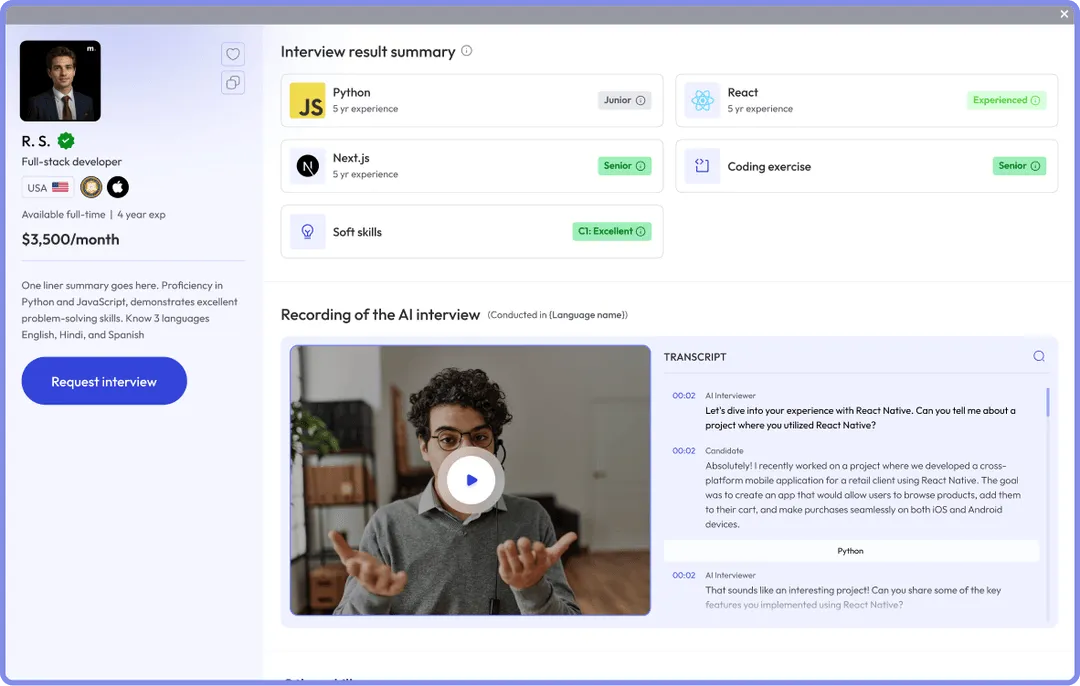}\\
    \footnotesize\textit{Note: Structured AI interview as seen by the candidate (left panel) and a sample interview report as seen by the recruiter (right panel).}
\end{figure}

Figure \ref{fig:screen_shots} presents a screen shot of a report generated from the AI interview.

\section{Experimental Balance Tables}

\paragraph{Experiment 1.} Randomization in Experiment 1 was stratified within job posting, so we report balance separately for the four roles. Table~\ref{tab:balance} presents SMDs on the AI skill ratings (observed for both arms because all Experiment 1 candidates completed the AI interview), demographics, and platform characteristics.

Most cells show $|\text{SMD}| < 0.25$. The exceptions cluster in the two smaller strata. In ML Engineer, treated candidates have systematically lower AI skill ratings (three of four skills show $|\text{SMD}| \geq 0.31$) along with lower Graduate Degree rates, Expected Hourly Rate, and Hours per Week Available, but higher LinkedIn referral rates. In Data Engineer and QA, treated candidates have higher Resume Scores, more Graduate Degrees, and more LinkedIn referrals, but the AI skill ratings are closely balanced. The imbalances are thus not aligned across covariate domains: where treatment looks stronger on paper credentials it does not generally look stronger on AI-assessed skill, and vice versa. No single covariate exceeds the 0.25 threshold in the same direction across all four roles. The pattern is consistent with chance variation in small cells rather than a systematic failure of randomization.

\begin{table}
\centering
\caption{Covariate Balance in Experiment 1 by Role: Treatment vs.\ Control}
\label{tab:balance}
\small
\resizebox{\textwidth}{!}{%
\begin{tabular}{l ccc ccc ccc ccc}
\toprule
& \multicolumn{3}{c}{\textbf{Frontend}} & \multicolumn{3}{c}{\textbf{QA}} & \multicolumn{3}{c}{\textbf{Data Eng.}} & \multicolumn{3}{c}{\textbf{ML Eng.}} \\
& \multicolumn{3}{c}{($N_C=266$, $N_T=265$)} & \multicolumn{3}{c}{($N_C=139$, $N_T=139$)} & \multicolumn{3}{c}{($N_C=95$, $N_T=96$)} & \multicolumn{3}{c}{($N_C=53$, $N_T=55$)} \\
\cmidrule(lr){2-4} \cmidrule(lr){5-7} \cmidrule(lr){8-10} \cmidrule(lr){11-13}
Variable & Control & Treated & SMD & Control & Treated & SMD & Control & Treated & SMD & Control & Treated & SMD \\
\midrule
\addlinespace[3pt]
\multicolumn{13}{l}{\textit{AI Interview Skill Ratings}} \\
\addlinespace[2pt]
Skill 1 (role-specific) & 1.38 & 1.38 & $\phantom{-}0.00$ & 1.60 & 1.52 & $-0.14$           & 1.75 & 1.75 & $\phantom{-}0.02$ & 1.68 & 1.48 & $\mathbf{-0.36}$ \\
Skill 2 (role-specific) & 1.29 & 1.20 & $-0.21$           & 1.60 & 1.62 & $\phantom{-}0.03$ & 1.63 & 1.73 & $\phantom{-}0.19$ & 1.69 & 1.51 & $\mathbf{-0.31}$ \\
Skill 3 (role-specific) & 1.50 & 1.40 & $-0.18$           & 1.47 & 1.36 & $-0.21$           & 1.65 & 1.61 & $-0.07$           & 1.68 & 1.43 & $\mathbf{-0.40}$ \\
Skill 4 (role-specific) & 1.30 & 1.24 & $-0.13$           & 1.43 & 1.24 & $\mathbf{-0.38}$  & 1.53 & 1.54 & $\phantom{-}0.01$ & 1.68 & 1.62 & $-0.09$ \\
\addlinespace[3pt]
\multicolumn{13}{l}{\textit{Demographics}} \\
\addlinespace[2pt]
Male                & 0.83  & 0.83  & $-0.00$           & 0.69  & 0.68  & $-0.01$           & 0.76  & 0.81  & $\phantom{-}0.11$          & 0.86  & 0.79  & $-0.18$ \\
Age                 & 26.60 & 26.14 & $-0.08$           & 29.61 & 30.06 & $\phantom{-}0.08$ & 27.44 & 26.95 & $-0.10$                    & 27.02 & 25.98 & $-0.19$ \\
Years of Experience & 5.07  & 4.47  & $-0.16$           & 6.44  & 7.73  & $\mathbf{0.26}$   & 5.12  & 5.04  & $-0.02$                    & 5.08  & 5.20  & $\phantom{-}0.03$ \\
College Degree      & 0.87  & 0.87  & $-0.02$           & 0.90  & 0.92  & $\phantom{-}0.09$ & 0.95  & 0.92  & $-0.14$                    & 0.94  & 0.97  & $\phantom{-}0.11$ \\
Graduate Degree     & 0.24  & 0.17  & $-0.17$           & 0.25  & 0.28  & $\phantom{-}0.08$ & 0.30  & 0.45  & $\mathbf{\phantom{-}0.32}$ & 0.55  & 0.38  & $\mathbf{-0.35}$ \\
\addlinespace[3pt]
\multicolumn{13}{l}{\textit{Platform \& Application Characteristics}} \\
\addlinespace[2pt]
Resume Score (0--100)      & 80.64 & 80.32 & $-0.02$           & 75.14 & 83.65 & $\mathbf{0.36}$   & 69.11 & 78.61 & $\mathbf{\phantom{-}0.46}$ & 74.81 & 79.27 & $\phantom{-}0.23$ \\
Notice Period (days)       & 9.44  & 10.27 & $\phantom{-}0.05$ & 8.81  & 13.80 & $\mathbf{0.33}$   & 16.57 & 17.61 & $\phantom{-}0.04$          & 8.08  & 7.57  & $-0.03$ \\
Hours per Week Available   & 35.44 & 35.44 & $-0.00$           & 37.12 & 38.29 & $\phantom{-}0.06$ & 36.93 & 37.44 & $\phantom{-}0.03$          & 40.17 & 34.21 & $\mathbf{-0.37}$ \\
Referred via LinkedIn      & 0.89  & 0.94  & $\phantom{-}0.18$ & 0.89  & 0.91  & $\phantom{-}0.06$ & 0.83  & 0.93  & $\mathbf{\phantom{-}0.32}$ & 0.73  & 0.89  & $\mathbf{\phantom{-}0.40}$ \\
\bottomrule
\end{tabular}
}
\vspace{4pt}
\begin{minipage}{\textwidth}
\footnotesize
\textit{Notes:} Standardized mean differences (SMD) use pooled standard deviations; bold indicates $|\text{SMD}| > 0.25$. AI skill ratings are on a 0--3 scale, where 0 = Not Familiar, 1 = Junior, 2 = Mid-level, and 3 = Senior. Role-specific skill definitions are as follows. \textit{Frontend}: 1 = React frontend, 2 = Core web tech, 3 = API integration, 4 = Testing/quality. \textit{QA}: 1 = Manual/automation testing, 2 = Test planning, 3 = QA tools, 4 = API/performance testing. \textit{Data Engineer}: 1 = Data pipelines, 2 = SQL/programming, 3 = Cloud platforms, 4 = Big data tools. \textit{ML Engineer}: 1 = Production ML, 2 = Python for ML, 3 = ML frameworks, 4 = MLOps.
\end{minipage}
\end{table}

\paragraph{Experiment 2.} Applicants invited to participate were randomized at application to the resume-only control arm and the AI-assisted treatment arm. Table~\ref{tab:summary_stats_treatment_balance} reports standardized mean differences (SMDs) on the pre-treatment covariates observed for both arms.

Groups are balanced. The largest absolute SMD is 0.04 (Resume Score); the remaining six covariates have $|\text{SMD}| \leq 0.04$. None of the covariates approach the 0.25 threshold that would warrant covariate-adjustment concern, consistent with the randomization producing arms that are statistically indistinguishable on observed background characteristics.

\begin{table}
\centering
\caption{Covariate Balance in Experiment 2: Treatment vs.\ Control}
\label{tab:summary_stats_treatment_balance}
\small
  \begin{tabular}{lccc}
  \toprule
  \multicolumn{4}{c}{Experiment 2 \quad ($N_C = 8{,}957$, $N_T = 25{,}536$)}\\
  \midrule
  Variable & Control & Treated & SMD \\
  \midrule
  \textit{Demographics} & & & \\
  Male & 0.77 & 0.77 & -0.01 \\
  Age & 25.10 & 24.96 & -0.03 \\
  Years of Experience & 3.48 & 3.47 & -0.00 \\
  High School & 0.01 & 0.01 & 0.01 \\
  Bachelor's & 0.52 & 0.53 & 0.01 \\
  Master's & 0.13 & 0.14 & 0.03 \\
  \textit{Platform \& Application Characteristics} & & & \\
  Resume Score (0--100) & 85.17 & 85.88 & 0.04 \\
  \bottomrule
  \end{tabular}
\vspace{4pt}
\begin{minipage}{\textwidth}
\footnotesize
\textit{Notes:} Standardized mean differences (SMD) use pooled standard deviations; bold indicates $|\text{SMD}| > 0.25$ (no covariates meet this threshold). Binary indicators are reported as proportions. AI-specific variables are observed only in the treatment group and are omitted from the balance comparison.
\end{minipage}
\end{table}

\section{Composition Effects of the AI-Assisted Pipeline}
\label{app:rank_hte}

This appendix supports the discussion in Section~\ref{sec:dropout} of how the AI-assisted pipeline reshapes the composition of advanced candidates relative to resume screening. We use the fact that we observe both an \emph{AI Score} and a \emph{Resume Score} for every treatment-arm completer. We rank candidates separately under each score and compute the change in rank when moving from \emph{Resume Score} to \emph{AI Score}. We then group candidates into terciles by the rank change. Candidates in the top tercile are advanced under the AI process more than they would be under resume screening (``Benefits from AI''); candidates in the bottom tercile are advanced less (``Harmed by AI'').

\begin{figure}[!htbp]
\centering
\caption{Average Covariate Values by Tercile of AI--Resume Rank Change}
\label{fig:rank_heatmap}
\includegraphics[scale=0.3]{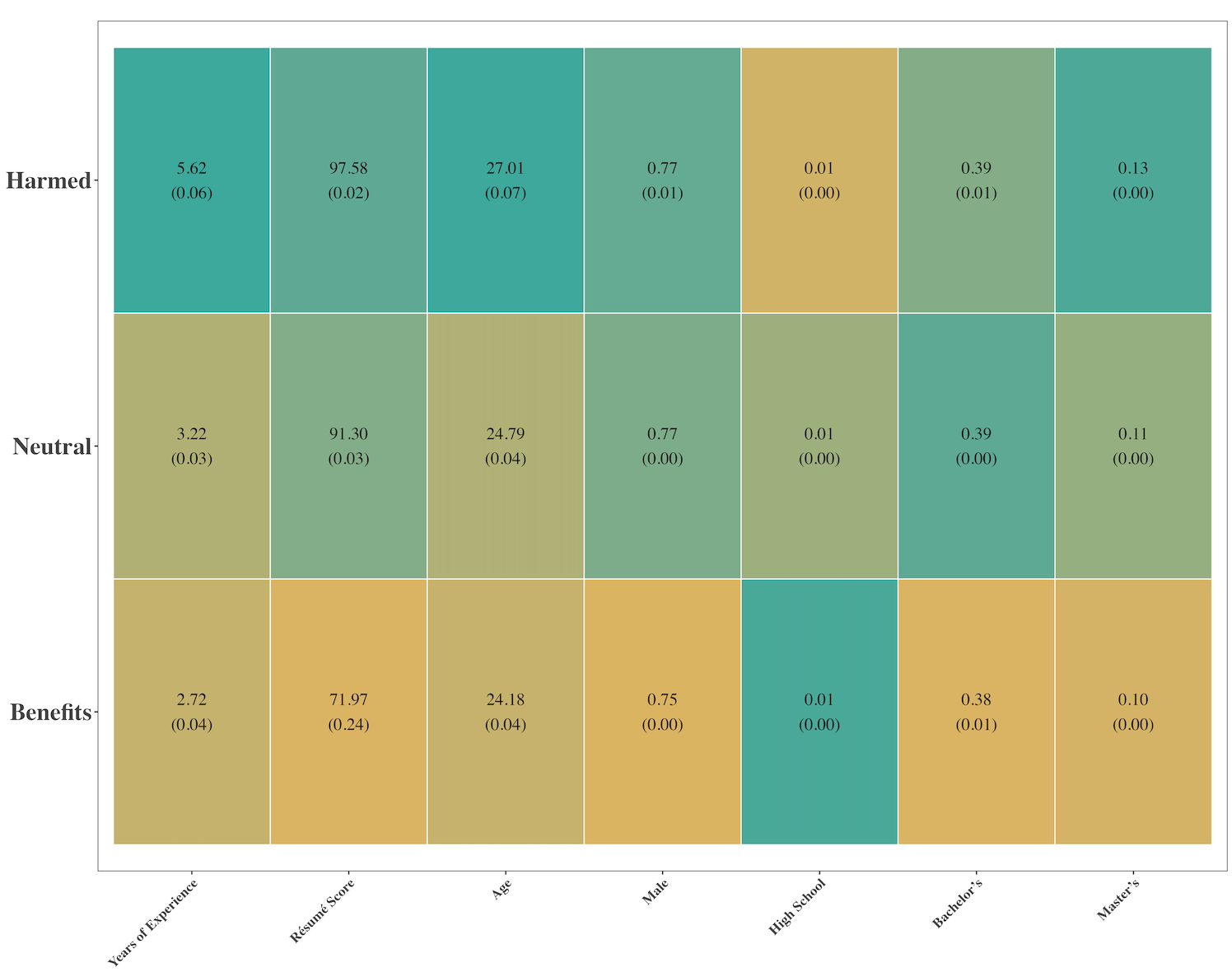}
\vspace{0.3em}
\begin{minipage}{\textwidth}
\footnotesize\textit{Notes:} Average covariate values across three equally sized groups, defined by the within-treatment-arm tercile of the change in candidate rank when moving from \emph{Resume Score} ranking to \emph{AI Score} ranking. ``Benefits from AI'' is the top tercile of rank change; ``Harmed by AI'' is the bottom tercile. Standard errors of the within-group means are reported in the heatmap.
\end{minipage}
\end{figure}

Figure~\ref{fig:rank_heatmap} displays average covariate values in the three terciles. Candidates who benefit from the AI process have, on average, 2.7 years of experience, against 5.6 in the bottom tercile; they are younger (24.2 vs.\ 27.0); they have substantially lower \emph{Resume Scores} (72.0 vs.\ 97.6); and they are less likely to hold a master's degree. The shift toward earlier-career candidates is consistent with Section~\ref{sec:auc}: the AI report's predictive lift is largest where conventional credentials are weakest, and the operational pipeline therefore moves shortlists toward exactly those candidates.

\section{Technical Details of Skill Extraction and Matching}
\label{app:technical_details}
Here, we provide the details of the extraction and matching process used to align self-reported skills with AI-vetted ratings. After normalization, each candidate's resume-listed skills were mapped to the AI-derived categories. Discrepancies were flagged when a candidate claimed a higher proficiency level than that assigned by the AI.

\subsection*{Pseudo-Code for Skill Extraction}

\begin{lstlisting}[label={lst:skill_matching_appendix}]
# Pseudocode for Skill Matching Functions

## 1. Function: `extract_junior_skills(skill_text)`
- **Input**: A string `skill_text` containing skills separated by '|'.
- **Process**:
  - Convert `skill_text` to lowercase.
  - Split the string by '|' and iterate through each skill.
  - Extract the first part of each skill (split by ':') and check if it contains the word "junior".
  - Return the list of skills that contain "junior".
- **Output**: A list of junior skills.

## 2. Function: `create_skill_matcher(model_name="gpt-4", temperature=0)`
- **Input**: Model name (`model_name`) and temperature for GPT-4 model.
- **Process**:
  - Create a GPT-4 model instance using the provided model name and temperature.

### Nested Function: `match_skill(skill, skill_list)`
- **Input**: A skill (`skill`) and a list of skills (`skill_list`).
- **Process**:
  - Generate a prompt that asks the model if the skill is present in the skill list.
  - Use the GPT model to check if the skill exists in the list, where the model responds with "True" or "False".
- **Output**: A response indicating whether the skill is present in the skill list (`True` or `False`).

- **Output**: The created model instance (`lm`).

## 3. Function: `process_skills_matching(df, skill_matcher)`
- **Input**: A dataframe `df` and the `skill_matcher` function.
- **Process**:
  - Iterate through each row in the dataframe.
  - Extract the junior skills from the `junior_skills` column using `extract_junior_skills()`.
  - Compare each junior skill with the skills in the `resume_skills` column using `skill_matcher()`.
  - Store the matched skills in a list.
- **Output**: The dataframe with a new column `matched_skills_junior` indicating the matched junior skills.
\end{lstlisting}
\captionof{lstlisting}{Pseudo-code demonstrating the LLM-based skill extraction and matching approach. The script parses skill data, standardizes variant names, and identifies discrepancies between resume claims and AI assessments.}

\section{Mechanical Pool-Size Null}
\label{sec:null_pool}

The +20~pp Experiment~2 treatment effect on final-interview pass rates could be order statistics. The treatment arm picks its top 35 from $N_T = 25{,}536$ completed applications; the control arm picks its top 35 from $N_C = 8{,}957$. Under the sharp null that the AI report carries no information beyond the platform's \emph{Resume Score}, drawing from the larger pool still raises selected quality mechanically. This appendix bounds that effect.

We draw latent applicant quality under two scenarios and select the top 35 in each arm. \emph{Scenario~A} samples quality from $\mathcal{N}(0,1)$, rescaled to the mean and standard deviation of the pooled \emph{Resume Score} distribution so that the slope below enters in fitted units. \emph{Scenario~B} bootstraps \emph{Resume Scores} directly from the empirical pool of 34,493 applicant resumes assembled from the two ranked applicant files. We run 5{,}000 bootstrap iterations per scenario.

We translate selected applicants' \emph{Resume Scores} into predicted pass rates via the logistic mapping $\Pr(\text{pass} \mid q) = \sigma(\alpha + \beta_{\text{resume}}(q - \mu_{RS}))$, where $\mu_{RS} = 85.66$ is the pooled Experiment-2 applicant \emph{Resume Score} mean (SD $\sigma_{RS} = 16.79$). The slope $\beta_{\text{resume}}$ comes from a logistic regression of final-interview pass on centered \emph{Resume Score}, fit on the 70 Experiment~2 finalists (35 treatment + 35 control) with no-shows coded as zero ($\hat\beta_{\text{resume}} = 0.049$).\footnote{The finalist sample is small; the slope is identified mainly off within-non-missing variation. It is sufficient to bound the order of magnitude of the mechanical effect.} We then solve for the intercept $\alpha$ in each scenario so that the bootstrap-averaged predicted control pass rate equals the observed $12/35 = 0.343$. Within each iteration the predicted pass rate is the mean of $\sigma(\alpha + \beta_{\text{resume}}(q - \mu_{RS}))$ over the 35 selected applicants; the bootstrap average then runs over the 5{,}000 iterations. This yields $\alpha_A = -3.155$ and $\alpha_B = -1.292$. By construction, control rates match in both scenarios. The treatment rate is what the treatment arm would generate if pool size were the only thing differing across arms.

\paragraph{Results.} Table~\ref{tab:null_pool} reports bootstrap means, standard errors, and 95\% confidence intervals. Under Scenario~A, the mechanical effect is $+4.5$~pp (95\% CI $[0.9, 8.1]$). Under Scenario~B, the empirical \emph{Resume Score} distribution has a hard ceiling: 12.5\% of applicants tie at the maximum (98.75). The top 35 in either arm almost always saturates at this value, and the mechanical difference collapses to exactly zero.

Order statistics alone explain at most a fifth of the effect.

\begin{table}[htbp]
\centering
\caption{Mechanical Pool-Size Null: Bootstrap Simulation}
\label{tab:null_pool}
\begin{threeparttable}
\begin{tabular}{lcc}
\toprule\toprule
Statistic & Scenario A (Normal) & Scenario B (Empirical RS) \\
\midrule
Mean quality, selected control & 136.61 $\pm$ 0.02 [134.31, 138.98] & 98.75 $\pm$ 0.00 [98.75, 98.75] \\
Mean quality, selected treatment & 140.66 $\pm$ 0.02 [138.57, 142.88] & 98.75 $\pm$ 0.00 [98.75, 98.75] \\
$\Delta$ quality (T$-$C) & 4.04 $\pm$ 0.02 [0.88, 7.16] & 0.00 $\pm$ 0.00 [0.00, 0.00] \\
\midrule
Predicted pass rate, control & 34.3\% $\pm$ 0.0\% [31.7, 36.9] & 34.3\% $\pm$ 0.0\% [34.3, 34.3] \\
Predicted pass rate, treatment & 38.8\% $\pm$ 0.0\% [36.4, 41.4] & 34.3\% $\pm$ 0.0\% [34.3, 34.3] \\
Mechanical $\Delta$ pass rate (T$-$C) & 4.5 $\pm$ 0.0 [0.9, 8.1] & 0.0 $\pm$ 0.0 [0.0, 0.0] \\
\midrule
Calibrated $\alpha$ & $-3.155$ & $-1.292$ \\
$N$ iterations & 5{,}000 & 5{,}000 \\
$N_C$ (control pool) & 8{,}957 & 8{,}957 \\
$N_T$ (treatment pool) & 25{,}543 & 25{,}543 \\
\bottomrule\bottomrule
\end{tabular}
\begin{tablenotes}[flushleft]
\footnotesize
\item \textit{Scenario A}: latent quality drawn i.i.d.\ $\mathcal{N}(\mu_{RS},\sigma_{RS}^2)$ where $\mu_{RS}$ and $\sigma_{RS}$ are the pooled Experiment-2 applicant Resume Score mean and SD.
\item \textit{Scenario B}: quality drawn by bootstrap resampling from the empirical pooled Experiment-2 applicant Resume Score distribution.
\item $\beta_{\text{resume}}$ fit via logistic regression on Experiment-2 finalists ($N=70$; 35 treatment + 35 control; control deduplicated by email and restricted to the first 35 by Interview Date; no-shows coded as fail). $\hat\beta_{\text{resume}} = 0.049$.
\item Calibration target: control predicted pass rate $= 12/35 = 0.343$ (no-shows-as-zero denominator). Top-$k = 35$ per arm. Observed difference in final-interview pass rates in Experiment-2: $7/35 \approx 0.200$.
\end{tablenotes}
\end{threeparttable}
\end{table}

\section{Additional Details on External Validation of AI Recruiter Scores}\label{audit_info}

\begin{table}[htbp]
\centering
\caption{Agreement between the external coder and original AI-generated labels by skill}
\label{tab:coder_agreement}
\begin{tabular}{lccccc}
\hline
Skill & Agreement & Wilson 95\% CI & $\kappa$ & 95\% CI for $\kappa$ & McNemar test \\
\hline
React & 60/60 = 100.0\% & [94.0\%, 100.0\%] & 1.00 & [1.00, 1.00] & No discordant pairs \\
JavaScript & 56/60 = 93.3\% & [84.1\%, 97.4\%] & 0.87 & [0.74, 0.99] & Exact $p = .125$ \\
HTML/CSS & 57/60 = 95.0\% & [86.3\%, 98.3\%] & 0.90 & [0.79, 1.00] & Exact $p = .250$ \\
\hline
All three skills matched & 54/60 = 90.0\% & [79.9\%, 95.3\%] & -- & -- & -- \\
\hline
\end{tabular}
\vspace{0.5em}
\begin{minipage}{0.95\textwidth}
\footnotesize
\textit{Note.} An independent external coder, unaffiliated with micro1 and blind to the AI-generated ratings, reviewed 60 AI interview transcripts from Experiment 2 and coded whether each candidate demonstrated React, JavaScript, and HTML/CSS skills. Because the original AI labels were ordinal whereas the external coding was binary, we collapsed the AI labels to experienced versus not experienced before comparing them. The table reports agreement rates, Wilson 95\% confidence intervals, Cohen's $\kappa$, and exact McNemar tests by skill.
\end{minipage}
\end{table}

\end{document}